\documentclass[10pt,journal,compsoc]{IEEEtran}

\ifCLASSOPTIONcompsoc
  \usepackage[nocompress]{cite}
\else
  \usepackage{cite}
\fi

\usepackage{amsmath}
\usepackage{mathtools}
\usepackage{amssymb}
\usepackage{amsthm}
\usepackage{physics}
\usepackage{siunitx}
\usepackage{bm}
\usepackage{bbm}
\usepackage{algorithm}
\usepackage{bbding}
\usepackage{algcompatible}
\usepackage{soul}
\usepackage{graphicx}
\usepackage{xcolor}
\usepackage{subfig}
\usepackage{changes}
\usepackage{tabularx}
\usepackage{multirow,ctable}
\makeatletter
\newtheorem{definition}{\textbf{Definition}}
\newtheorem{lemma}{\textbf{Lemma}}
\newtheorem{theorem}{\textbf{Theorem}}

\newtheorem{assumption}{\textbf{Assumption}}
\theoremstyle{definition}
\newtheorem{remark}{\textbf{Remark}}

\newcounter{relctr} %% <- counter for relations
\everydisplay\expandafter{\the\everydisplay\setcounter{relctr}{0}} %% <- reset every eq
\newcommand\labelrel[2]{%
  \begingroup
    \refstepcounter{relctr}%
    \stackrel{\textnormal{(\alph{relctr})}}{\mathstrut{#1}}%
    \originallabel{#2}%
  \endgroup
}
\AtBeginDocument{\let\originallabel\label} %% <- store original definition

\ifCLASSINFOpdf
\else
\fi

\begin{document}
\title{DP$^2$-FedSAM: Enhancing Differentially Private Federated Learning Through Personalized Sharpness-Aware Minimization}
\author{Zhenxiao~Zhang,~\IEEEmembership{Student~Member,~IEEE,} Yuanxiong~Guo,~\IEEEmembership{Senior~Member,~IEEE,} and Yanmin~Gong,~\IEEEmembership{Senior~Member,~IEEE}
\IEEEcompsocitemizethanks{\IEEEcompsocthanksitem Z. Zhang, and Y. Gong are with the Department of Electrical and Computer Engineering, The University of Texas at San Antonio, San Antonio, TX, 78249. Y. Guo is with the Department of Information Systems and Cyber Security, The University of Texas at San Antonio, San Antonio, TX, 78249. E-mail: \{zhenxiao.zhang@my., yuanxiong.guo@, yanmin.gong@\}utsa.edu.}
% \IEEEcompsocitemizethanks{\IEEEcompsocthanksitem Z. Zhang and Z. Gao both contributed equally to this work.}
}

\IEEEtitleabstractindextext{%
\begin{abstract}
Federated learning (FL) is a distributed machine learning approach that allows multiple clients to collaboratively train a model without sharing their raw data. To prevent sensitive information from being inferred through the model updates shared in FL, differentially private federated learning (DPFL) has been proposed. DPFL ensures formal and rigorous privacy protection in FL by clipping and adding random noise to the shared model updates. However, the existing DPFL methods often result in severe model utility degradation, especially in settings with data heterogeneity. To enhance model utility, we propose a novel DPFL method named DP$^2$-FedSAM: Differentially Private and Personalized Federated Learning with Sharpness-Aware Minimization. DP$^2$-FedSAM leverages personalized partial model-sharing and sharpness-aware minimization optimizer to mitigate the adverse impact of noise addition and clipping, thereby significantly improving model utility without sacrificing privacy. From a theoretical perspective, we provide a rigorous theoretical analysis of the privacy and convergence guarantees of our proposed method. To evaluate the effectiveness of DP$^2$-FedSAM, we conduct extensive evaluations based on common benchmark datasets. Our results verify that our method improves the privacy-utility trade-off compared to the existing DPFL methods, particularly in heterogeneous data settings.
\end{abstract}

}

\maketitle

\IEEEdisplaynontitleabstractindextext
\IEEEpeerreviewmaketitle

\IEEEraisesectionheading{\section{Introduction}\label{sec:introduction}}
Federated Learning (FL) is a machine learning paradigm where multiple clients collaboratively learn a shared model without sharing their training datasets. In FL, each client trains the model locally on their data and only shares the model updates with a central server. The server aggregates these updates to improve the global model and sends the updated global model back to the clients for further training. While this paradigm significantly enhances data privacy and reduces the need for data centralization, it is insufficient to guarantee data privacy. An adversary can still recover the private data using reconstruction attack~\cite{zhu2019deep} or infer whether a sample is in the training dataset using membership inference attack~\cite{salem2019ml} by observing the model updates from a client.

To address the privacy issues, differential privacy (DP) has been integrated into FL to provide a formal and strong privacy guarantee. Client-level DP in FL was first introduced in \cite{McMahan2018learning} to protect the privacy of all examples contributed by a client during the training process. As FedAvg is the most common FL algorithm, DP-FedAvg is a natural choice to provide client-level guarantee in FL. In general, DP-FedAvg clips the local updates by a threshold and then adds the Gaussian noise with magnitude proportional to the threshold to the clipped local updates.

Although DP-FedAvg can provide a rigorous client-level DP guarantee, it faces challenges in maintaining high model accuracy due to the clipping and noise addition operations. To overcome these challenges, existing studies have proposed methods such as restricting the norm of local updates~\cite{cheng2022differentially}, leveraging sparsification techniques~\cite{cheng2022differentially,hu2022federated}, and utilizing flat landscape optimization~\cite{shi2023make} to mitigate the adverse effects of clipping and noise addition. However, under heterogeneous data distributions in FL, the performance of these methods is still limited. Specifically, restricting the norm of local updates can reduce the impact of noise but may compromise the model's accuracy; sparsification techniques might lead to accuracy instability in the presence of imbalanced data; and flat landscape optimization can enhance the local model's robustness, but its global flatness cannot be guaranteed in significantly heterogeneous data distributions.

\begin{figure}[t]
  \centering
  \includegraphics[width=0.38\textwidth]{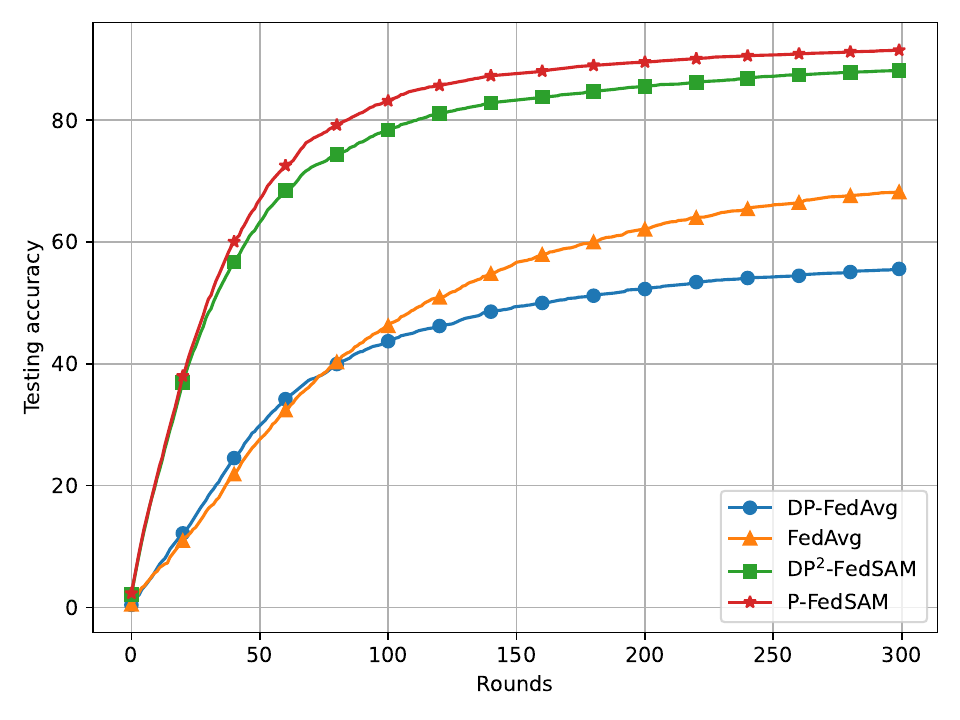}
   \caption{Test accuracy on CIFAR-10 with a CNN for different methods under a non-IID data partition, where 1000 clients each has data from only 2 classes. P-FedSAM is essentially DP$^2$-FedSAM without the mechanisms of clipping and adding noise. DP$^2$-FedSAM exhibits enhanced robustness compared to DP-FedAvg. }
   \label{fig_compare}
\end{figure}

In this paper, we propose a simple yet powerful framework called \emph{DP$^2$-FedSAM}: \textbf{D}ifferentially \textbf{P}rivate and \textbf{P}ersonalized \textbf{Fed}erated \textbf{L}earning with \textbf{S}harpness-\textbf{A}ware \textbf{M}inimization. DP$^2$-FedSAM leverages the partial model personalization and sharpness-aware local training in FL to reduce the adverse impacts of clipping and noise addition and improve model utility without sacrificing privacy. As shown in Fig.~\ref{fig_compare}, our proposed method is more robust in the private setting under data heterogeneity than DP-FedAvg. For instance, DP$^2$-FedSAM exhibits a modest decrease of approximately {$4\%$} in test accuracy in the private setting on the CIFAR-10 dataset with a CNN model under non-IID data distributions, whereas DP-FedAvg experiences a more substantial drop of around {$13\%$}. The benefits behind this phenomenon can be attributed to the following three aspects: 1) We minimize the norm of local updates among heterogeneous clients by partial model personalization. That is, instead of training a shared full model with high inconsistency across clients under heterogeneous data distributions, we train a single shared representation extractor while enabling each client to have a personalized classifier head. Consequently, our method can reduce the bias introduced by clipping in DP training. 2) We use sharpness-aware training to generate local flat models. These flat models exhibit smaller variations with respect to parameter changes, leading to smaller norm of local updates to reduce the clipping error. 3) By combining partial-model sharing and sharpness-aware training, we can obtain a global flat model after aggregation even in the heterogeneous data setting. This global flat minimum demonstrates greater resilience compared to its sharp counterpart under the same noise magnitude in DP training. Moreover, in the region near flat minima, the model is better positioned to follow an accurate gradient descent path, resulting in faster convergence.

In summary, the main contributions of this paper can be summarized as follows:
\begin{itemize}
    \item We propose a novel DPFL scheme named DP$^2$-FedSAM, which utilizes partial model personalization and sharpness-aware minimization to improve model utility without sacrificing privacy under data heterogeneity.
    
    \item We provide rigorous theoretical analysis on both the convergence and privacy guarantees of DP$^2$-FedSAM.
    
    \item Extensive evaluations based on common benchmark datasets verify our proposed scheme could improve the privacy-utility trade-off compared with the state-of-the-art methods in DPFL. 
\end{itemize}

The rest of this paper is organized as follows. Preliminaries on FL and DP are described in Section~\ref{sec:preliminary}. Section~\ref{sec:methodology} presents the proposed DP$^2$-FedSAM scheme. The convergence and privacy properties of DP$^2$-FedSAM are rigorously analyzed in Section~\ref{sec:theo_ana}. Section~\ref{sec:exp_results} shows the experimental results. Section~\ref{sec:related} reviews related work. Finally, Section~\ref{sec:conclusion} concludes the paper. 
%%%%%%%%

%%%%%%%%%%%%%%

%-------------------------------------------------
\section{Preliminaries}\label{sec:preliminary}
%-------------------------------------------------
%----------------------------------------
\begin{table}[t]
  \caption{Summary of main notations.}
  \label{tab:notations}
  \centering
  \begin{tabular}{cc}
    \toprule
    Notation & Definition\\
    \midrule
    $i, j$ &  Index for client\\
    $N$ & Total number of clients\\
    $[N]$ & \{1, 2, \ldots, $N$\}\\
    $r$ & Client sampling ratio\\
    $t$ & Index for global iteration\\
    $\mathcal{S}^t$ & Set of selected clients in iteration $t$\\
    ${D}_i$ & Local dataset of client $i$\\
    $F_i(\cdot)$ & Local objective function of client $i$ \\
    $\phi$ & Shared representation extractor\\
    $h_i$ & Personal classifier of client $i$\\
    $s$ & Index for local iteration\\
    $t$ & Index for communication round\\
    $\tau_h$ & Total number of local iterations for $h$\\
    $\tau_{\phi}$ & Total number of local iterations for $\phi$\\        
    $C$ & Clipping threshold \\
    $\sigma$ & Noise multiplier \\
    $\Delta_i^t$ & Model updates\\
    $\tilde{\Delta}_i^t$ & Clipped model updates\\
    $\hat{\Delta}_i^t$ & Noisy model updates\\
    $\eta_h$ & Learning rate for $h$\\
    $\eta_{\phi}$ & Learning rate for $\phi$\\
    $p$ & Perturbation of SAM \\
    $q$ & Perturbation parameter\\
    $\sigma_{\phi}$, $\sigma_h$ & Bounded variances \\
   $\epsilon$, $\delta$, $\alpha$, $\rho$ & Differential privacy parameters\\
  \bottomrule
\end{tabular}
\end{table}

%%%%%%%%%%%%%%%%%%%%%%%%%%%%%%%%%%%%%%%
\subsection{Federated Learning}
%%%%%%%%%%%%%%%%%%%%%%%%%%%%%%%%%%%%%%%

We consider a FL system that consists of $N$ clients and a server to collaboratively solve the following optimization problem:
\begin{equation}\label{pro_fl}
\min_{{\theta} \in \mathbb{R}^d} F({\theta}) := \frac{1}{N} \sum_{i=1}^N F_i({\theta}),
\end{equation}
where $F_i({\theta}) := (1/|D_{i}|)|\sum_{\xi \in D_{i}} l_i({\theta}; \xi)$ is the local objective function of client $i$, and $D_i$ is the local dataset of client $i$. Here $l_i$ is the loss function defined by the learning task, and $\xi$ represents a data sample from $D_i$. A list of main notations used in the paper is summarized in Table~\ref{tab:notations}.

%%%%%%%%%%%%%%%%%%%%%%%%%%%%%%%%%%%%%%%
\subsection{SAM}\label{subsec:sam}
%%%%%%%%%%%%%%%%%%%%%%%%%%%%%%%%%%%%%%%
The goal of SAM~\cite{foret2021sharpness} is to seek out model parameter values whose entire neighborhoods have uniformly low training loss values, thereby leveraging the flatness geometry of the loss landscape to improve model generalization ability. This can be achieved by solving the min-max problem:
\begin{equation}\label{prob:sam_obj}
\min_{\theta} \max_{\norm{p}_2 \leq q} F(\theta + p),
\end{equation}
where $q$ is a predefined constant controlling the radius of the perturbation $p$. Given the difficulty of precisely identifying the optimal direction $p^*=\arg\max_{\norm{p}_2 \leq q}F(\theta + p)$, SAM approximately solves it via the use of the first-order Taylor expansion of $F$. Specifically, SAM updates the model weights ${\theta}$ in two steps. First, it computes the stochastic gradient $\widetilde{\nabla}_{ \theta} F( {\theta})$ and calculates the perturbation $p^*$ as follows:
\begin{equation}
      p^* = q\frac{\widetilde{\nabla}_{ {\theta}}F( {\theta})}{\norm{\widetilde{\nabla}_{ {\theta}}F( {\theta})}_2}.
\end{equation}
Then the perturbation is used to update the parameters as follows:
\begin{align}
{\theta}= {\theta}-\eta_{\theta}\widetilde{\nabla}_{ {\theta}}F( {\theta}+p^*),
\end{align}
where $\eta_{\theta}$ is the learning rate.

%%%%%%%%%%%%%%%%%%%%%%%%%%%%%%%%%%%%%%
\subsection{Differential Privacy}
%%%%%%%%%%%%%%%%%%%%%%%%%%%%%%%%%%%%%%

DP provides a rigorous notion to prevent privacy leakage and has become the de-facto standard for measuring privacy risk \cite{dwork2014algorithmic}. In this paper, we consider client-level DP, which ensures the adversary cannot distinguish whether a target client is present in the dataset or not. The formal client-level DP is defined as follows. 

%%%------------------------------------------------------------
\begin{definition}[Client-level $(\epsilon,\delta)$-DP~\cite{McMahan2018learning}]\label{def:dp}
Given privacy parameters $\epsilon>0$ and $0\leq\delta<1$, a random mechanism $\mathcal{M}$ satisfies $(\epsilon,\delta)$-DP if for any two neighboring datasets $D, D^\prime$ constructed by adding or removing all records of a client, and any subset of outputs $\mathcal{O} \subseteq range(\mathcal{M})$, we have
\begin{align}
\textup{Pr}[\mathcal{M}(D)\in \mathcal{O}]\leq e^{\epsilon}\textup{Pr}[\mathcal{M}(D^\prime)\in \mathcal{O}]+\delta.
\end{align}
When $\delta = 0$, we have $\epsilon$-DP. 
\end{definition}
%%%-------------------------------------------------------------
A smaller parameter $\epsilon$ provides a stronger privacy guarantee but typically results in a lower utility. The parameter $\delta$ is usually set to a small value to account for the probability that the inequality fails. Client-level DP aims to protect the privacy of any client's participation from the aggregated model update. Therefore, it is essential to ensure that local updates remain similar, whether or not a client chooses to participate. 

To better quantify the privacy loss across multiple iterations in differentially private learning algorithms, we consider R\'enyi DP (RDP)~\cite{mironov2017renyi}, which is a relaxed version of $(\epsilon, \delta)$-DP. It is defined as follows.
% -------------------------------------------%
\begin{definition}[$(\alpha,\rho)$-RDP~\cite{mironov2017renyi}]
\label{def:rdp}
Given a real number $\alpha>0$ and privacy parameter $\rho\geq0$, a random mechanism $\mathcal{M}$ satisfies $(\alpha,\rho)$-RDP if for any two neighboring datasets $D, D^\prime$ that differs in one client's records, the R\'enyi $\alpha$-divergence between $\mathcal{M}(D)$ and $\mathcal{M}(D^\prime)$ satisfies
\begin{equation}
    D_{\alpha}[\mathcal{M}(D)\|\mathcal{M}(D^\prime)]\coloneqq \frac{1}{\alpha-1}\log \mathbb{E}\left[\left(\frac{\mathcal{M}(D)}{\mathcal{M}(D^\prime)}\right)^{\alpha}\right]\leq\rho.
\end{equation} 
\end{definition}
% --------------------------------------------%
The Gaussian mechanism is commonly used to achieve $(\epsilon,\delta)$-DP by injecting zero-mean Gaussian noise to the query output, the scale of which depends on the $\ell_2$-sensitivity of the query function. The definition of $\ell_2$-sensitivity is given as follows.\\
\begin{definition}[$\ell_2$-sensitivity~\cite{dwork2014algorithmic}]
\label{def:sensitivity}
Let $f:\mathcal{D} \rightarrow \mathbb{R}^d$ be a query function over a dataset. The $\ell_2$-sensitivity of $f$ is defined as
\begin{equation}
\psi(f)\coloneqq\max_{D \simeq D^\prime }\|f(D)-f(D^\prime)\|_2
\end{equation} where $D$ and $D^\prime$ are two neighboring datasets. 
\end{definition}
% -------------------------------------------%

In the following, we provide some useful lemmas about DP and RDP that will be used to derive the main results of this paper.
% %-------------------------------------------%

% %-------------------------------------------%
\begin{lemma}[Gaussian Mechanism \cite{mironov2017renyi}]\label{lemma:gaussian_mechanism}
Let $f: \mathcal{D} \rightarrow \mathbb{R}^d$ be a query function with $\ell_2$-sensitivity $\psi(h)$. The Gaussian mechanism $\mathcal{M} = f(D) + \mathcal{N}(0, \sigma^2 \psi(f)^2 \bm{I}_d)$ satisfies $(\alpha,\alpha /2\sigma^2)$-RDP.
\end{lemma}
% -------------------------------------------%
%-------------------------------------------%
\begin{lemma}[From RDP to $(\epsilon, \delta)$-DP \cite{wang2019subsampled}]\label{lemma:rdp_dp}
If the randomized mechanism $\mathcal{M}$ satisfies $(\alpha, \rho(\alpha))$-RDP, then it also satisfies $(\rho(\alpha) + \frac{\log(1/\delta)}{\alpha-1}, \delta)$-DP.
\end{lemma}
%-------------------------------------------%

%-------------------------------------------%
\begin{lemma}[RDP Composition \cite{mironov2017renyi}]\label{lemma:rdp_comr} 
For randomized mechanisms $\mathcal{M}_1$ and $\mathcal{M}_2$ applied on dataset $D$, if $\mathcal{M}_1$ satisfies $(\alpha, \rho_1)$-RDP and $\mathcal{M}_2$ satisfies $(\alpha, \rho_2)$-RDP, then their composition $ \mathcal{M}_1 \circ \mathcal{M}_2$ satisfies $(\alpha, \rho_1+\rho_2)$-RDP.
\end{lemma}
%-------------------------------------------%
In DP mechanisms, the privacy amplification property of DP allows for improved privacy guarantees without the need to increase the amount of added noise. Specifically, by applying a DP mechanism to a random subset of a dataset, it can achieve stronger privacy protection compared to applying it to the entire dataset. This concept, known as privacy amplification by subsampling, formally enhances the privacy guarantees of DP algorithms. The formal statement of privacy amplification by subsampling is given as follows:
% -------------------------------------------%
\begin{lemma}[RDP for Subsampling Mechanism \cite{wang2019subsampled}]\label{lemma:rdp_sub}
For a Gaussian mechanism $\mathcal{M}$ and any m-datapoints dataset $D$, define $\mathcal{M}\circ \textit{SUBSAMPLE}$ as 1) subsample without replacement $B$ data points from the dataset (denote $r=B/m$ as the sampling ratio); and 2) apply $\mathcal{M}$ on the subsampled dataset as input. Then if $\mathcal{M}$ satisfies $(\alpha,\rho(\alpha))$-RDP with respect to the subsampled dataset for all integers $\alpha \geq 2$, then the new randomized mechanism $\mathcal{M}\circ \textit{SUBSAMPLE}$ satisfies $(\alpha,\rho^\prime(\alpha))$-RDP w.r.t $D$, where
\begin{multline*}
\rho^\prime(\alpha) \leq \frac{1}{\alpha-1}\log \bigg(1 + r^2 {\binom{\alpha}{2}}\min\{4(e^{\rho(2)}-1), 2e^{\rho(2)}\}\\ 
+ \sum_{l=3}^{\alpha}r^l\binom{\alpha}{l}2e^{(l-1)\rho(l)}\bigg).
\end{multline*}
If $\sigma^2 \geq 0.7$ and $\alpha \leq (2/3)\sigma^2\psi^2(h) \log(1/q\alpha(1+\sigma^2)) + 1$, $\mathcal{M}\circ \textit{SUBSAMPLE}$ satisfies $(\alpha, 3.5 q^2 \alpha / \sigma^2)$-RDP. 
\end{lemma}
%-------------------------------------------%

%%%%%%%%%%%%%%%%%%%%%%%%%%%%%%%%%%%
\subsection{Attack Model and Privacy Goal}
%%%%%%%%%%%%%%%%%%%%%%%%%%%%%%%%%%%
In this paper, we consider the server to be ``honest-but-curious''. This means that the server is curious about a specific client's local dataset and intends to infer information from the shared messages, while honestly following the protocols involving the training process. Additionally, there may be a third party, such as an external observer or an unauthorized participant, that can intercept and analyze the global model broadcasted by the server at the end of each round. The privacy objective of this paper is to ensure that neither the server nor the third party can gain significant insights into a client's local dataset by observing the received global model update in each round.

%%%%%%%%%%%%%%%%%%%%%%%%%%%%%%%%%%
\subsection{DP-FedAvg: Achieving Client-level DP in FL}
%%%%%%%%%%%%%%%%%%%%%%%%%%%%%%%%%%
%%%%%%%%%%%%%%%%%%%%%%%%%%%%%%%%%%%%%%%
\begin{algorithm}[t]  
\textbf{Input:} Initial server model ${\theta}^0$, aggregation period $\tau$, total rounds $T$, sample size $r$, clipping threshold $C$, noise magnitude $\sigma$, and learning rate $\eta$.\\
\textbf{Output:} Final global model ${\theta}^{T}$ 
\caption{DP-FedAvg~\cite{McMahan2018learning} }\label{algorithm-dpfedavg}
\begin{algorithmic}[1]
    %  Initialization:  
    \FOR{ $t = 0, \ldots, T - 1$}
        \STATE Uniformly sample a set $\mathcal{S}^t\subseteq[N]$ with $r=|\mathcal{S}^t|$ \label{dpfl_server_sample}
        \STATE Broadcast ${\theta}^t$ to all clients in $\mathcal{S}^t$\label{dpfl_server_broad}
        \FOR{each client $i \in \mathcal{S}^t$ \textbf{in parallel}}
            \STATE ${\theta}_i^{t,0}\gets$ ${\theta}^t$\label{dpfl_client_init}  \FOR{$s=0,\dots,\tau-1$}\label{dpfl_client_sgd_start}
                \STATE Compute a mini-batch stochastic gradient $\mathbf{g}_{i}^{t,s-1}$
            \STATE ${\theta}_{i}^{t,s}\gets$ ${\theta}_{i}^{t,s-1}-\eta\mathbf{g}_{i}^{t,s-1}$
        \ENDFOR\label{dpfl_client_sgd_end}
        \STATE $\Delta_{i}^{t}= {\theta}_i^{t,\tau} - {\theta}^{t}$\label{dpfl_client_cal_upd}
        \STATE $\tilde{\Delta}_{i}^{t} = \Delta_{i}^{t}/\max\left(1,\frac{\norm{\Delta_{i}^{t}}_2}{C}\right)+\mathcal{N}(0,\frac{C^2\sigma^2\mathbf{I}_d}{r})$\label{dpfl_client_clip_noise}
    \ENDFOR
    \STATE ${\theta}^{t + 1} \gets {\theta}^t + \frac{1}{r}\sum_{i\in\mathcal{S}^t}\tilde{\Delta}_{i}^{t}$\label{dpfl_server_aggr}
\ENDFOR
\end{algorithmic}
\end{algorithm}
%%%%%%%%%%%%%%%%%%%%%%%%%%%%%%%%%%%%%%%
To provide client-level DP in FL under an ``honest-but-curious'' server, DP can be adapted to this setting by perturbing the model updates locally before uploading them to the server. Specifically, as shown in Algorithm~\ref{algorithm-dpfedavg}, DP-FedAvg consists of the following steps in each FL round $t$. 1) Server sends the global model to a randomly sampled subset of clients (lines~\ref{dpfl_server_sample}-\ref{dpfl_server_broad}). 2) Each client initializes its local model to be the received global model (line~\ref{dpfl_client_init}), performs $\tau$ steps of SGD  (lines~\ref{dpfl_client_sgd_start}-\ref{dpfl_client_sgd_end}) and computes its local model update (line~\ref{dpfl_client_cal_upd});
3) Each client clips the norm of model updates $\Delta_{i}^{t}$ by a threshold $C$ and adds Gaussian noise to its bounded local model update (line~\ref{dpfl_client_clip_noise}); 
4) Server aggregates the perturbed local model updates received from the clients to update the global model (line~\ref{dpfl_server_aggr}).

Although DP-FedAvg ensures client-level DP, the utility of the resulting global model is significantly diminished due to the clipping and noise addition operations. Specifically, Clipping model updates, which limits sensitivity to individual data points, may restrict the model's convergence by limiting updates towards the dominant gradient. Additionally, the addition of Gaussian noise, intended to enhance privacy, can introduce bias into the learning process, thereby risking suboptimal convergence and potentially degrading the overall model performance. This motivates us to develop a new DPFL framework that can maintain high utility while ensuring client-level DP.

%%%%%%%%%%%%%%%%%%%%%%%%%%%%%%%%%%%%%%
\section{Methodology}\label{sec:methodology}
%%%%%%%%%%%%%%%%%%%%%%%%%%%%%%%%%%%%%%

In this section, we first analyze the impact of the clipping operation and introduce partial model-sharing and SAM to mitigate its effects. While both methods can effectively alleviate the impact of clipping, they fall short in reducing noise-induced errors individually under data heterogeneity. To address this, we combine SAM with partial model-sharing to achieve a globally flatter minimum, thereby making the model more robust against the error introduced by adding noise.

\subsection{The Impact of Clipping}

We start by analyzing the impact of clipping operation in DP-FedAvg. We denote $\norm{\cdot}$ as the $\ell_2$ vector norm and define the clipping operation as $\text{clip}(\Delta_i^t, C)=\Delta_{i}^{t}/\max\left(1,{\norm{\Delta_{i}^{t}}_2}/{C}\right)$. The clipping operation ensures that the norm of $\Delta_i^t$ does not exceed the threshold $C$. The error between local model updates before and after clipping can be expressed as follows:
\begin{equation}\label{eq_clipping}
    \|\Delta_i^t - \text{clip}(\Delta_i^t, C)\| = 
\begin{cases} 
\|\Delta_i^t\| - C & \text{if } \|\Delta_i^t\| > C, \\
0 & \text{otherwise}.
\end{cases}
\end{equation}
\eqref{eq_clipping} indicates that {\bf reducing the norm of local updates} can effectively reduce the error introduced by clipping. However, the norm of local updates remains large, especially in data heterogeneity settings. To address this issue, we propose using partial model-sharing and SAM to reduce the norm of local updates, thereby mitigating the clipping error.
%%%%%%%%%%%%%%%%%%%%%%%%%
\subsection{Partial Model-Sharing Mitigates the Effect of Clipping}\label{subsec:partial_clipping}
%%%%%%%%%%%%%%%%%%%%%%%%%

In order to achieve a better privacy-utility trade-off, empirical evidence suggests that a smaller clipping threshold is preferable as it effectively mitigates the substantial variance resulting from the injected noise~\cite{de2022unlocking}. However, in the standard FL scenario with heterogeneous data distributions, local updates remain notably large even when approaching the global optimal solution. Therefore, using a small clipping threshold will introduce a substantial and persistent bias, as shown in Equation~\eqref{eq_clipping}. 

Motivated by the observation that clients in FL tend to have minimal discrepancies in their data representations while displaying substantial differences in their classifier heads~\cite{bengio2013representation,chen2020simple,collins2021exploiting}, we propose a personalized FL strategy with partial model-sharing. This approach involves training a single shared private representation extractor while allowing each client to maintain a personalized classifier head. This approach helps to reduce the norm of local updates caused by data heterogeneity because the shared representation extractors are approximately homogeneous, thereby mitigating the error caused by the clipping in the data heterogeneity settings.

Therefore, an effective alternative to Equation~\eqref{pro_fl} is to train a shared representation extractor among all clients while allowing each client to personalize its model through a customized classifier head. Formally, we consider the model $\theta\in \mathbb{R}^d$ to be divided into two parts: global shared representation $\phi\in\mathbb{R}^{d_1}$ and personal classifier head $h\in\mathbb{R}^{d_2}$ with $d = d_1+d_2$. Under the above notion, our goal is to solve the optimization problem:
\begin{equation}\label{obj_partial_model_share}
\min_{ {\phi}, \{ {h}_i\}_{i=1}^{N}} \frac{1}{N}\sum_{i=1}^{N} F_{i}( {\phi}, {h}_i),
\end{equation}
where $\phi$ is the representation extractor shared by all clients, and $h_i$ is the local personalized classifier head for client $i$.

%%%%%%%%%%%%%%%%%%%%%%%%%
\subsection{SAM Mitigates the Effect of Clipping}\label{subsec:sam_clipping}
%%%%%%%%%%%%%%%%%%%%%%%%%
In the region near flat minima, the loss function exhibits smaller variations with respect to parameter changes, resulting in the smaller norm of model updates~\cite{park2023differentially}.  This inherent property is especially important in the context of data heterogeneity, as it reduces the probability of the norms of model updates exceeding the clipping threshold, thereby decreasing the frequency of clipping occurrences and ultimately reducing the error introduced by clipping. Moreover, in the region near flat minima, the model is better positioned to follow an accurate gradient descent path, resulting in faster convergence.

Inspired by the advantages brought by flat minima and partial-model sharing, we propose to use SAM training in the update of shared representation extractors, resulting in smaller norms of model updates and reducing the clipping error. Specifically, we jointly minimize the local loss function for client $i$ and smooth its loss landscape by solving the following optimization problem: 
\begin{equation}
\min_{\phi} \max_{\norm{p}_2 \leq q} F_i (\phi + p, h_i^{t+1}),
\end{equation}
where $q$ is a predefined constant controlling the radius of the perturbation $p$. Following the standard steps in Section~\ref{subsec:sam}, SAM updates the representation extractor ${\phi}_{i}$ in two steps. First, it computes the partial stochastic gradient $\widetilde{\nabla}_{ {\phi}} F_i( {\phi}_{i}^{t,s}, {h}_{i}^{t+1})$ and calculates the perturbation $p( {\phi}_{i}^{t,s})$ as follows:
\begin{equation}\label{alg_cal_per}
      p( {\phi}_{i}^{t,s}) = q\frac{\widetilde{\nabla}_{ {\phi}}F_i( {\phi}_{i}^{t,s}, {h}_{i}^{t+1})}{\norm{\widetilde{\nabla}_{ {\phi}}F_i( {\phi}_{i}^{t,s}, {h}_{i}^{t+1})}}.
\end{equation}
Then the perturbation is used to update the shared parameters as follows:
\begin{align}\label{alg_phi_upd}
{\phi}_{i}^{t,s+1}= {\phi}_{i}^{t,s}-\eta_{\phi}\widetilde{\nabla}_{ {\phi}}F_i( {\phi}_{i}^{t,s}+p( {\phi}_{i}^{t,s}), {h}_{i}^{t+1}),
\end{align}
where $\eta_{\phi}$ is the learning rate of shared representation extractor.  
%%%%%%%%%%%%%%%%%%%%%%%%%%%%%%%%%%%%
\begin{figure}[t]
  \centering
   \includegraphics[width=0.38\textwidth]{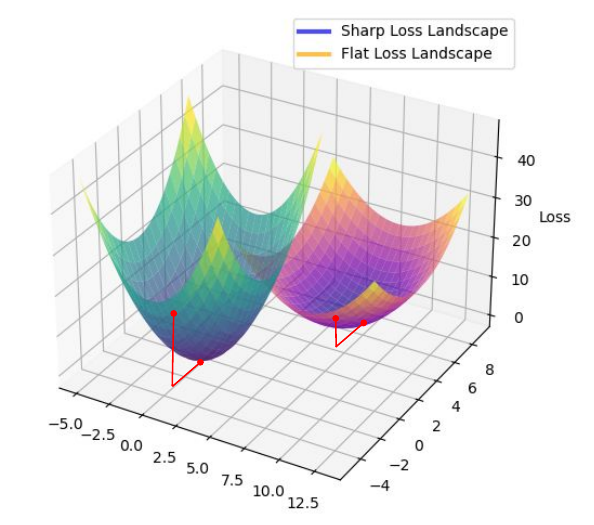}
   \caption{Illustration of sharp and flat loss landscape. The flat minimum is more robust than the sharp one under the same perturbation in DP training.}
   \label{fig:land}
\end{figure}
%%%%%%%%%%%%%%%%%%%%%%%%%%%%%%%%%%%% 

%%%%%%%%%%%%%%%%%%%%%%%%%
\subsection{Mitigating the Effect of Adding Noise by Combining Partial Model-Sharing and SAM}
%%%%%%%%%%%%%%%%%%%%%%%%%
Sections~\ref{subsec:partial_clipping} and \ref{subsec:sam_clipping} have pointed out how both partial-model sharing and SAM can reduce clipping error, but there remains a crucial step in DPFL: the addition of noise. Here, we first point out that either method alone is insufficient to mitigate the errors introduced by adding noise. Then, we find that combining partial model-sharing and SAM can effectively reduce the noise introduced by the noise addition operation. 

\begin{remark}{\bf Partial model-sharing alone is insufficient to mitigate the effect of adding noise.}\label{remark:partial_alone}
Partial model-sharing can reduce the error introduced by clipping by introducing approximate homogeneous shared representation extractors in heterogeneous data distributions. However, the shared representation extractor may be highly sensitive to the bias introduced by adding noise. This sensitivity can lead to reduced robustness against noise-induced errors, ultimately impacting overall model performance.
\end{remark}

\begin{remark}{\bf SAM alone is insufficient to mitigate the effect of adding noise.}\label{remark:sam_alone}
Although SAM can reduce the clipping error by seeking local minima and thereby reducing the norm of local model updates, local flat models do not necessarily lead to an aggregated global flat model in FL due to data heterogeneity. This discrepancy can diminish the model's robustness against noise-induced errors, ultimately affecting the overall performance.
\end{remark}

In terms of noise addition, as shown in Fig.~\ref{fig:land}, a sharp minimum is more sensitive to the perturbation introduced by the same additive noise than a flat minimum. Therefore, our goal is to provide an aggregated global flat minimum to enhance robustness against noise in DPFL. As discussed in Remark~\ref{remark:sam_alone}, local flat minima do not necessarily lead to a global flat model due to data heterogeneity. However, by combining partial model-sharing and SAM, we can achieve a global flat model even in data heterogeneity settings.
\begin{remark}{\bf Combining partial model-sharing and SAM is sufficient to mitigate the effect of adding noise.}
      Local flat minima do not necessarily result in a globally flat model due to data heterogeneity. However, partial model-sharing can provide more consistent shared representation extractors. By sharing these homogeneous representation extractors, even in the presence of data heterogeneity, we can achieve a global flat minimum. 
\end{remark}

%%%%%%%%%%%%%%%%%%%%%%%%%%%%%%%%%%%%%%%%%%%%%%%%%%%%%%%%%%%%
\subsection{DP$^2$-FedSAM Algorithm}
%%%%%%%%%%%%%%%%%%%%%%%%%%%%%%%%%%%%%%%%%%%%%%%%%%%%%%%%
%%%%
\begin{figure*}
  \centering
   \includegraphics[width=0.83\linewidth]{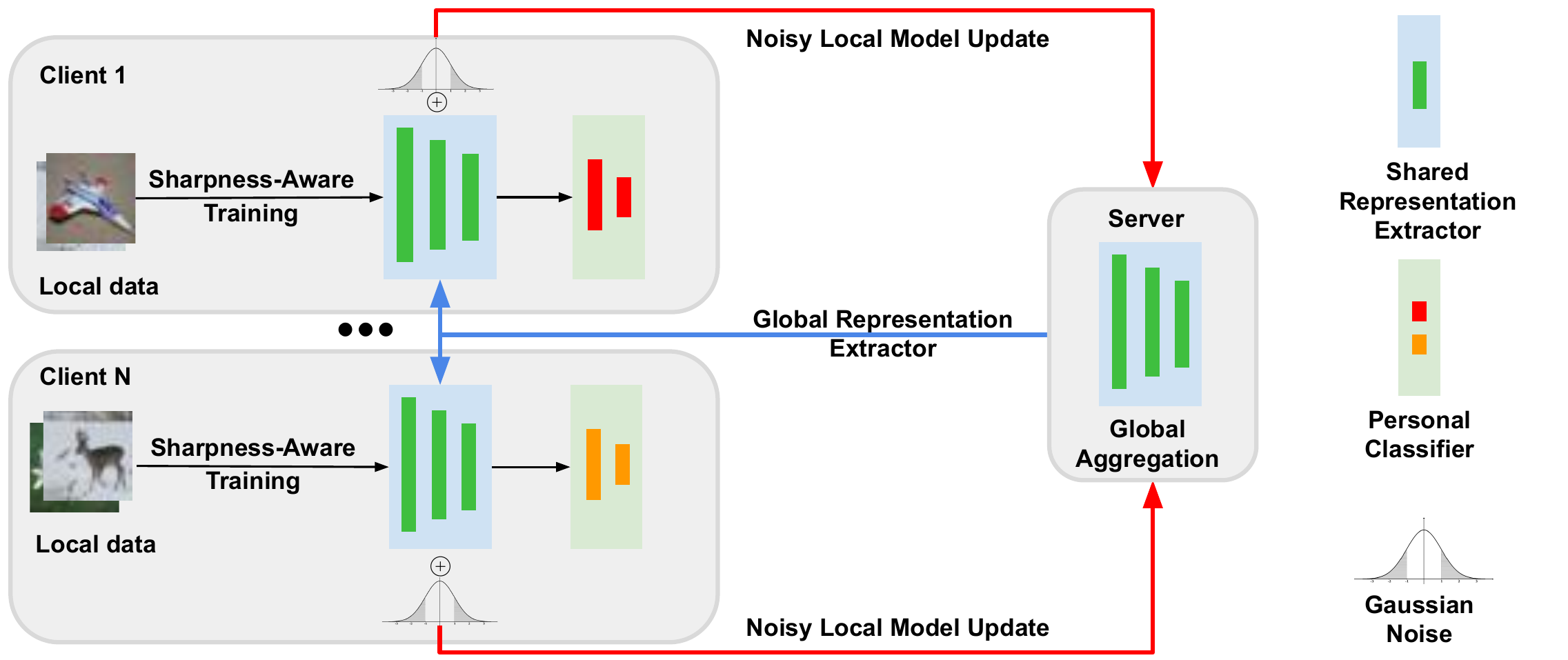}
    \caption{An overview of DP$^2$-FedSAM. Partial model personalization allows each client to locally retain a personal classifier and only share the representation extractor with the server for aggregation. The server aggregates the shared representation extractor and sends it back to all clients. During local training, SAM is applied to enhance the robustness of the shared representation extractor.}\label{fig_pfedsam}
\end{figure*}

Inspired by the advantages of combining partial model personalization and SAM training to alleviate the impacts of clipping and noise addition, we propose DP$^2$-FedSAM to mitigate the model utility degradation in DPFL by strategically integrating the aforementioned two modules. The overview of DP$^2$-FedSAM is shown in Fig.~\ref{fig_pfedsam}. Intuitively, both SAM training and partial model-sharing generate local updates with small norms to reduce the clipping error, and integrating them can provide a flatter global model, offering better stability and perturbation resilience in data heterogeneity settings. 

The pseudo-code for the proposed DP$^2$-FedSAM is provided in Algorithm~\ref{alg_dp2fedsam}. At each round $t$, the server uniformly and randomly samples a set $\mathcal{S}^t$ of $rN$ clients (line~\ref{alg_sample}). Then, the current global version of the shared partial model $\phi^t$ is broadcast to selected clients (line~\ref{alg_broadcast}). After that, each client performs $\tau_h$ steps of SGD to update its personal classifier head $h_i$ while keeping the received shared parameters $\phi^t$ fixed (lines~\ref{alg_head_upd_start}-\ref{alg_head_upd_end}). Next, the shared representation extractor is updated by $\tau_{\phi}$ steps of the SAM optimizer, with the new personal classifier fixed (lines~\ref{alg_initial_2}-\ref{alg_rep_upd_end}). Specifically, each client first calculates the gradient perturbation (line~\ref{alg_cal_pertur}) using the stochastic gradient (line~\ref{alg_cal_stochastic}), and then updates the shared representation extractor (line~\ref{alg_rep_update}). Then, each client $i$ updates the shared representation part (line~\ref{alg_upd_share}) and calculates the model updates $\Delta_i^t$ (line~\ref{client_cal_upd}). Since there is no a priori bound on the model updates, each client first clips its local model updates by a threshold $C$ such that $\norm{\Delta_i^t}_2\leq C$ (line~\ref{alg_clip}). Next, each client perturbs its clipped model updates by adding independent Gaussian noise $\mathcal{N}(0,{C^2\sigma^2\mathbf{I}_{d_1}}/{rN})$, where $\sigma$ is the noise multiplier (line~\ref{alg_add_noise}). Then, the noisy local updates $\hat{\Delta}_{i}^{t}$ are uploaded to server (line~\ref{alg_upload}). For the unselected clients, their local personalized classifiers remain unchanged (lines~\ref{alg_keep_start}-\ref{alg_keep_end}). Finally, the server uses the estimated aggregated model updates to update the global shared representation extractor for the next round (line~\ref{alg_aggr}).

%%%%%%%%%%%%%%%%%%%%%%%%%%%%%%%%
\begin{algorithm}[t]
    \caption{DP$^{2}$-FedSAM}\label{alg_dp2fedsam}
    \textbf{Input:} Initial states $ {\phi}^0, \{{h}_{i}^{0}\}_{i = 1}^{N}$, client sampling ratio $r$, number of local iterations $\tau_h$, $\tau_{\phi}$, number of communication rounds $T$, learning rates $\eta_h$, $\eta_{\phi}$, and neighborhood size $q$ \\
    \textbf{Output:} Personalized models $( {\phi}^{T},  {h}_i^{T}),\forall i\in[N].$ 
    \begin{algorithmic}[1]
    \FOR {$t = 0, 1, \dots, T-1$}
        \STATE Server randomly samples a set of $rN$ clients $\mathcal{S}^t $.\label{alg_sample}
        \STATE Server broadcasts the current global version of the shared parameters $ {\phi}^t$ to all clients in $\mathcal{S}^t$.\label{alg_broadcast}
        \FOR{each client $i \in \mathcal{S}^t$ \text{in parallel}}
            \STATE Initialize ${h}_i^{t,0} = {h}_i^t$ \label{alg_initial}
            \FOR{$s=0,\dots,\tau_h-1$}\label{alg_head_upd_start}
                \STATE Compute stochastic gradient                $\widetilde{\nabla}_{{h}}F_i({\phi}^t, {h}_{i}^{t,s})$
                \STATE $h_{i}^{t,s+1}= h_{i}^{t,s} -\eta_{h}\widetilde{\nabla}_{h}F_i({\phi}^t, {h}_{i}^{t,s})$\label{alg_h_upd}
            \ENDFOR\label{alg_head_upd_end}
            \STATE Update ${h}_{i}^{t+1}= {h}_{i}^{t, \tau_h}$ and initialize ${\phi}_i^{t,0} = {\phi}^t$  \label{alg_initial_2}
            \FOR {$s = 0, \dots, \tau_{\phi}-1$}\label{alg_rep_upd_start}
                \STATE Compute stochastic gradient $\widetilde{\nabla}_{\phi}F_i({\phi}^t, {h}_{i}^{t+1})$ \label{alg_cal_stochastic}
                \STATE Gradient perturbation by Equation~\eqref{alg_cal_per} \label{alg_cal_pertur}
                % $p( {\phi}_{i}^{t,s}) = q\frac{\widetilde{\nabla}_{ {\phi}}F_i( {\phi}_{i}^{t,s}, {h}_{i}^{t+1})}{\norm{\widetilde{\nabla}_{ {\phi}}F_i( {\phi}_{i}^{t,s}, {h}_{i}^{t+1})}_2}$
                \STATE Local representation update by Equation~\eqref{alg_phi_upd} \label{alg_rep_update}
            \ENDFOR \label{alg_rep_upd_end}
            \STATE Update ${\phi}_i^{t+1} =  {\phi}_i^{t,\tau_{\phi}}$ \label{alg_upd_share}
            \STATE {$\Delta_{i}^{t}= {\phi}_i^{t+1} - {\phi}_i^{t}$}\label{client_cal_upd}
            \STATE $\tilde{\Delta}_{i}^{t} = \Delta_{i}^{t}\cdot\min\left(1,\frac{C}{\norm{\Delta_{i}^{t}}_2}\right)$\label{alg_clip}
            \STATE $\hat{\Delta}_{i}^{t} = \tilde{\Delta}_{i}^{t}+\mathcal{N}(0,\frac{C^2\sigma^2\mathbf{I}_{d_1}}{rN})$\label{alg_add_noise}
            % \State $\Delta c_i = c - \frac{\hat{\Delta}_i^t}{\eta_{\phi}\tau_{\phi}}$
            % \State $c_i = c_i + \Delta c_i$
            \STATE Client sends $\hat{\Delta}_{i}^{t}$ back to server \label{alg_upload}
        \ENDFOR
        \FOR{each client $i \notin \mathcal{S}^t$}\label{alg_keep_start}
            \STATE $ {h}_{i}^{t+1} = {h}_{i}^{t}$
        \ENDFOR  \label{alg_keep_end}

    \STATE Server updates ${\phi}^{t+1} =  {\phi}^{t} + \frac{1}{rN}\sum_{i\in\mathcal{S}^{t}} \hat{\Delta}_{i}^{t}$\label{alg_aggr}
    % \State $c = c+\frac{1}{rN}\sum_{i\in\mathcal{S}^{t}}\Delta c_i$
    \ENDFOR
    \end{algorithmic}
\end{algorithm}
%%%%%%%%%%%%%%%%%%%%%%%%%%%%%%%%%%%%%%%%%%%%%%%%%%%%%%%%%%%%%%

%%%%%%%%%%%%%%%%%%%%%%%%%%%%%%%%%%%%
\section{Theoretical Analysis}\label{sec:theo_ana}
%%%%%%%%%%%%%%%%%%%%%%%%%%%%%%%%%%%%
In this section, we provide the convergence results and end-to-end privacy guarantee of DP$^2$-FedSAM. Due to the page limit, we present the main theorems in this section and only provide proof sketches, leaving the complete proofs in the appendix. Before stating our theoretical results, we make the following assumptions for the convergence analysis.
 \begin{assumption}
[Smoothness]\label{assum_smooth}
For each $i\in[N]$, the function $F_i$ is continuously differentiable. There exist constants $L_{\phi}$, $L_{h}$, $L_{{\phi h}}$, $L_{{h\phi}}$ such that for each $i\in[N]$: 
\begin{itemize}
    \item $\nabla_{ {\phi}}F_{i}( {\phi}, {h_i})$ is $L_{{\phi}}$-Lipschitz with respect to $\phi$ and $L_{\phi h}$-Lipschitz with respect to $h_i$, and 
    \item $\nabla_{h} F_{i}(\phi,h_i)$ is $L_{h}$-Lipschitz with respect to $h_i$ and $L_{h\phi }$-Lipschitz with respect to $\phi$.
\end{itemize}
The relative cross-sensitivity of $\nabla_{\phi}F_i$ with respect to $h_i$ and $\nabla_{h}F_i$ with respect to $\phi$ is defined by the following scalar: 
\begin{equation}
    \chi \coloneqq \max\{L_{\phi h}, L_{h\phi}\}/\sqrt{L_{\phi}L_{h}}.
\end{equation}    
\end{assumption}

\begin{assumption}[Bounded Variance]\label{assum_bound_var}
    The stochastic gradients in Algorithm~\ref{alg_dp2fedsam} are unbiased and have bounded variance. That is, for all ${\phi}$ and $h_i$,
    \begin{align}
        \mathbb{E}[\widetilde{\nabla}_{ {\phi}}F_i(\phi, h_i)] = \nabla_{ {\phi}}F_i(\phi, {h}_{i}),\\
        \mathbb{E}[\widetilde{\nabla}_{h}F_i(\phi, {h}_{i})] = \nabla_{ {h}}F_i( {\phi}, {h}_{i}).
     \end{align}
    Furthermore, there exist constants $\sigma_{\phi}$ and $\sigma_{h}$ such that
    \begin{align}
        \mathbb{E}\norm{\widetilde{\nabla}_{ {\phi}}F_i(\phi, {h}_{i}) - \nabla_{{\phi}}F_i({\phi}, {h}_{i})}_2^2 \leq \sigma_{\phi}^2,\\
        \mathbb{E}\norm{\widetilde{\nabla}_{{h}} F_i( {\phi}, {h}_{i}) - \nabla_{{h}}F_i({\phi}, {h}_{i})}_2^2 \leq \sigma_{h}^2.
    \end{align}
\end{assumption}
%-----------------------------
\begin{assumption}[Bounded Gradient]\label{assum_bound_grad} For any $i\in[N]$, $\phi\in\mathbb{R}^{d_1}$, and $h_i\in\mathbb{R}^{d_2}$, we have $\norm{\nabla F_{i}(\phi,h)}_2\leq G$.
\end{assumption}
%-----------------------------
Assumptions~\ref{assum_smooth}--\ref{assum_bound_grad} are standard in the analysis of the convergence of FL~\cite{pillutla2022federated,bottou2018optimization,guo2022hybrid,shi2023make,zhang2024scalable}. For ease of notation, we denote $\Delta F_0 = F( {\phi}^0, {H}^0)-F^*$ with $F^*$ being the minimal value of $F(\cdot)$. Further, we use the shorthands ${H}^t = ( {h}_1^t,\cdots, {h}_N^t)$, $F(\phi,H)=1/N\sum_{i=1}^{N} F({\phi}^t, {h}_i^t)$, $\nabla_{\phi}^t=\norm{\nabla_{ {\phi}} F( {\phi}^t, {H}^t)}_2^2$, and $\nabla_{h}^t=1/N\sum_{i=1}^{N}\norm{\nabla_{h} F( {\phi}^t, {h}_i^t)}_2^2$.

%%%%%%%%%%%%%%%%%%%%%%%%%%%%%%
\subsection{Convergence Analysis}
%%%%%%%%%%%%%%%%%%%%%%%%%%%%%%
In this subsection, we provide the convergence result of DP$^2$-FedSAM under the non-convex and non-IID setting in Theorem~\ref{theo:conv}. Before stating the final result, we highlight the unique challenges of our setting. 
%%%%%%%%%%%%%%
\subsubsection{Technical Challenges}
%%%%%%%%%%%%%%
The first challenge is that in DP$^2$-FedSAM, unlike the traditional FL where a single global model is shared among all clients, each client maintains a personalized model $h_i$. Thus, directly applying the analysis used for the shared global model would overlook the effects of these local personalized models, resulting in a loose bound. To address this issue, our key idea is to build the convergence analysis for both the global part $\nabla_{\phi}$ and the personalized part $\nabla_{h}$, so that we can achieve a more accurate bound that properly incorporates the influence of the personalized models.

The second challenge involves dealing with dependent random variables. Consider the iterates $(\phi^t, H^t)$ generated by DP$^2$-FedSAM. To analyze the effect of the $\phi$-update, the smoothness of $F(\cdot, H^t)$ is utilized as follows:
\begin{multline}\label{eq_smooth}
    F(\phi^{t+1}, H^{t+1})- F(\phi^{t}, H^{t})\leq\langle\nabla_{\phi}F(\phi^t,H^{t+1}), \phi^{t+1}-\phi^t\rangle \\
     +\frac{L_{\phi}}{2}\norm{\phi^{t+1}-\phi^t}.
\end{multline}
For the standard convergence proofs of stochastic gradient methods, simplification is achieved on the first term on RHS of~\eqref{eq_smooth} when taking the expectation of $\mathcal{S}^t$, as the gradient is usually independent of $\mathcal{S}^t$. However, this is not the case of DP$^2$-FedSAM. Specifically, 
\begin{multline}
% \end{multline}
   \mathbb{E}_t[ \langle\nabla_{\phi}F(\phi^t,H^{t+1}), \phi^{t+1}-\phi^t\rangle] \\\neq  \langle\mathbb{E}_t[\nabla_{\phi}F(\phi^t,H^{t+1})], \mathbb{E}_t[\phi^{t+1}-\phi^t]\rangle,
\end{multline}
where $\mathbb{E}_t$ denotes the expectation w.r.t. $\mathcal{S}^t$. This discrepancy arises because $H^{t+1}$ is already updated based on $\mathcal{S}^t$, making both $H^{t+1}$ and $\phi^{t+1}$ dependent random variables due to their mutual dependence on the sampling $\mathcal{S}^t$. Therefore, directly taking expectation w.r.t. $\mathcal{S}^t$ does not yield a useful result. We introduce virtual full participation to decouple the dependent random variables to overcome this challenge. Define $\check{H}^{t+1}$ as the result of local $h$-updates as if all clients had participated. This iterate is virtual and it is a tool of the analysis but is not required by the algorithm. Since $\check{H}^{t+1}$ is deterministic, we can now take an expectation w.r.t. the sampling $\mathcal{S}^t$ over $\phi^{t+1}$ only, then we can simplify the inner product term as 
\begin{multline}
    \mathbb{E}_t[ \langle\nabla_{\phi}F(\phi^t,\check{H}^{t+1}), \phi^{t+1}-\phi^t\rangle] \\=  \langle\mathbb\nabla_{\phi}F(\phi^t,\check{H}^{t+1}), \mathbb{E}_t[\phi^{t+1}-\phi^t]\rangle.
\end{multline}
We refer to Appendix~\ref{appen:proof_conv} for more details.

%%%%%%%%%%%%%%
\subsubsection{Convergence of DP$^2$-FedSAM}
%%%%%%%%%%%%%%

In this subsection, we propose our main convergence results of the proposed DP$^2$-FedSAM algorithm in the following theorem. We only provide the proof sketch here and include the detailed proofs in the appendices.

\begin{lemma}[Convergence Decomposition]\label{lemma_decomp}. Under Assumption~\ref{assum_smooth}, we have
\begin{multline*}
    \mathbb{E}_t[F( {\phi}^{t+1}, {H}^{t+1}) - F( {\phi}^{t}, {H}^{t})]\\ \leq \underbrace{\langle \nabla_{\phi}F( {\phi}^{t}, \check{H}^{t+1}),  {\phi}^{t+1}- {\phi}^{t}\rangle}_{\mathcal{T}_{1,\phi}} + \underbrace{L_{{\phi}}\mathbb{E}_t\norm{ {\phi}^{t+1}- {\phi}^{t}}^2}_{\mathcal{T}_{2,\phi}}\\ + \underbrace{\mathbb{E}_t[F( {\phi}^{t}, {H}^{t+1})- F( {\phi}^{t}, {H}^{t})]}_{\mathcal{T}_{{1, h}}}+ \underbrace{\frac{\chi^2 L_{h}}{2n}\sum_{i=1}^{n}\norm{\check{h}_i^{t+1}-h_i^{t+1}}}_{\mathcal{T}_{2, h}} .
\end{multline*}
\end{lemma}
\begin{IEEEproof}
    We start with 
\begin{align*}
    \mathbb{E}_t[F( {\phi}^{t+1},& {H}^{t+1}) - F( {\phi}^{t}, {H}^{t})] \\
     & = \underbrace{\mathbb{E}_t[F( {\phi}^{t+1}, {H}^{t+1}) - F( {\phi}^{t}, {H}^{t+1})]}_{\mathcal{T}_{{\phi}}} \\
    &\quad+ \underbrace{\mathbb{E}_t[F( {\phi}^{t}, {H}^{t+1})- F( {\phi}^{t}, {H}^{t})]}_{\mathcal{T}_{{1,h}}}
\end{align*}
For $\mathcal{T}_{{\phi}}$, we have
\begin{align*}
    \mathcal{T}_{{\phi}} &\labelrel\leq{t_phi_smooth} \langle \nabla_{\phi} F( {\phi}^{t}, {H}^{t+1}),  {\phi}^{t+1}- {\phi}^{t}\rangle + \frac{L_{{\phi}}}{2}\mathbb{E}_t\norm{ {\phi}^{t+1}- {\phi}^{t}}^2\\
    &= \langle \nabla_{\phi}F(\phi^t,H^{t+1}) - \nabla_{\phi}F(\phi^t,\check{H}^{t+1}), \phi^{t+1}-\phi^t \rangle \\
    & \quad+ \langle \nabla_{\phi}F( {\phi}^{t}, \check{H}^{t+1}),  {\phi}^{t+1}- {\phi}^{t}\rangle + \frac{L_{{\phi}}}{2}\mathbb{E}_t\norm{ {\phi}^{t+1}- {\phi}^{t}}^2\\
    &\labelrel\leq{t_phi_smooth1}\langle \nabla_{\phi}F( {\phi}^{t}, \check{H}^{t+1}),  {\phi}^{t+1}- {\phi}^{t}\rangle + L_{{\phi}}\mathbb{E}_t\norm{ {\phi}^{t+1}- {\phi}^{t}}^2 \\
    &\quad + \frac{1}{2L_{\phi}}\norm{\nabla_{\phi}F(\phi^t,H^{t+1}) - \nabla_{\phi}F(\phi^t,\check{H}^{t+1})}^2\\
    & \labelrel\leq{t_phi_smooth3} \underbrace{\langle \nabla_{\phi}F( {\phi}^{t}, \check{H}^{t+1}),  {\phi}^{t+1}- {\phi}^{t}\rangle}_{\mathcal{T}_{1,\phi}} + \underbrace{L_{{\phi}}\mathbb{E}_t\norm{ {\phi}^{t+1}- {\phi}^{t}}^2}_{\mathcal{T}_{2,\phi}} \\
    & \quad+ \underbrace{\frac{\chi^2 L_{h}}{2n}\sum_{i=1}^{n}\norm{\check{h}_i^{t+1}-h_i^{t+1}}}_{\mathcal{T}_{2,h}}
\end{align*}
where (\ref{t_phi_smooth}) and (\ref{t_phi_smooth3}) follow from Assumption~\ref{assum_smooth} and (\ref{t_phi_smooth1}) follows from the inequality that $2\langle \bm{a},\bm{b}\rangle \leq \gamma\norm{\bm{a}}^2 +\gamma^{-1}\norm{\bm{b}}^2, \forall \gamma \geq 0, \bm{a},\bm{b}\in \mathbb{R}^d.$
\end{IEEEproof}

Lemma~\ref{lemma_decomp} provides the decomposition of the total convergence error. By conducting a detailed analysis of the bounds for the decomposed terms $\mathcal{T}_{1,\phi},\mathcal{T}_{2,\phi}, \mathcal{T}_{1,h}, \mathcal{T}_{2,h}$ (see the details in Appendix~\ref{appen:proof_conv}, we will integrate these bounds into Lemma~\ref{lemma_decomp} to determine the overall convergence results of DP$^2$-FedSAM. 
 
\begin{theorem}[Convergence of Algorithm~\ref{alg_dp2fedsam}]\label{theo:conv}Under Assumptions~\ref{assum_smooth}-\ref{assum_bound_grad}, if the learning rates satisfy $\eta_{\phi} = \mathcal{O}(1/(\tau_{\phi}L_{\phi}\sqrt{T}))$, $\eta_{h} = \mathcal{O}(1/(\tau_{h}L_{h}\sqrt{T}))$, we have  

\begin{align}\label{eq_theo}
    \frac{1}{T}&\sum_{t=0}^{T-1}(\frac{\bar{\alpha}^t}{L_{\phi}}\mathbb{E}[\nabla_{\phi}^t]+\frac{r}{L_{h}}\mathbb{E}[\nabla_{h}^t])\leq\frac{\Delta F_0}{\sqrt{T}} \nonumber \\+ &\mathcal{O}\big(\eta_{\phi}^3\frac{1}{T}\sum_{t=0}^{T-1}{\bar{\alpha}^t}(G^2+\sigma_{\phi}^2)\big) + \mathcal{O}\big(\eta_{h}^2\sigma_h^2\big)\nonumber \\
    & +\mathcal{O}\big(\eta_{\phi}\frac{1}{T}\sum_{t=0}^{T-1}{\tilde{\alpha}^t}q^2\big) + \mathcal{O}(\frac{\sigma^2C^2d_1^2}{\eta_{\phi}r^2N^2}),
\end{align}
where 
\begin{equation}
    \bar{\alpha}^t = \frac{1}{N}\sum_{i=1}^{N}\alpha_i^t \quad\textit{and} \quad \tilde{\alpha}^t=\frac{1}{N}\sum_{i=1}^{N}|\alpha_i^t-\bar{\alpha}^t|,
\end{equation}
with $\alpha_i^t = \min(1,\frac{C}{\eta_{\phi}\norm{\sum_{s=0}^{\tau_{\phi}-1}\widetilde{\nabla}_{\phi}F_i^{t,s}(i)}_2})$, respectively. 
\end{theorem}

\begin{IEEEproof}
The proof is given in Appendix~\ref{appen:proof_conv}.
\end{IEEEproof}
\begin{remark}
    We can see that the convergence bound in Theorem~\ref{theo:conv} contains five parts. The first three terms are the same as the optimization error bound in FL with partial model personalization~\cite{pillutla2022federated}. The fourth term is the SAM perturbation error~\cite{shi2023make} because it is directly related to the SAM perturbation radius $q$. When $q$ is proportional to the learning rate, e.g., $q=\mathcal{O}(1/\sqrt{T})$, this term will vanish as the communication rounds $T$ increase. The last term is the privacy error~\cite{hu2022federated}.  When there is no privacy noise, i.e., $\sigma=0$, the privacy error is equal to zero.
\end{remark}
%%%%%%%%%%%%%%%%%%%%%%%%%%%%%%%%%%
\subsection{Privacy Analysis}
%%%%%%%%%%%%%%%%%%%%%%%%%%%%%%%%%%

% \red{(We should include the convergence analysis first.)}

Before stating our rigorous privacy analysis, we first provide the sensitivity analysis of the aggregated local updates in the clipping and noise addition operations. Assume two neighboring sets $\mathcal{S}$ and $\mathcal{S}^\prime$ differ in one client index $i^\prime$ such that $\mathcal{S}^\prime = \mathcal{S}^t\cup\{i^\prime\}$ or $\mathcal{S}^\prime = \mathcal{S}^t\setminus\{i^\prime\}$. For any adjacent datasets $D\coloneqq\{D_i\}_{i\in\mathcal{S}}$ and $D^\prime\coloneqq\{D_j\}_{j\in\mathcal{S}^\prime}$, according to Definition~\ref{def:sensitivity}, we have the following results.

%-----------------------------------------------
\begin{lemma}[Sensitivity]\label{lemma_sens}
    The $\ell_2$-sensitivity of the sum of local model updates is $C$.
\end{lemma}
%-----------------------------------------------
\begin{proof} For any adjacent datasets $D$ and $D^\prime$, the $\ell_2$-sensitivity of the sum of local model updates is
\begin{equation}
    \max_{D \simeq D^\prime }\Bigg\|\sum_{i\in\mathcal{S}}\Delta_i^t -\sum_{j\in\mathcal{S}^\prime}\Delta_i^t\Bigg\|_2 = \norm{\Delta_{i^\prime}^t}_2.
\end{equation}
Due to the clipping, we have the $\norm{\Delta_{i^\prime}^t}_2\leq C$. Thus, we have the final result.  
\end{proof}
After clipping and adding Gaussian noise, we provide the end-to-end privacy analysis of DP$^2$-FedSAM as follows.
%-----------------------------------------------
\begin{theorem}[Privacy Guarantee of DP$^2$-FedSAM]\label{theo_priv}
    Suppose clients are sampled without replacement with probability $r$ at each round. For any $\epsilon < 2\log(1/\delta)$ and $\delta\in(0,1)$, DP$^2$-FedSAM satisfied $(\epsilon, \delta)$-DP after $T$ communication rounds if
\begin{equation*}
\sigma^2 \geq \frac{7r^2T(\epsilon + 2\log(1/\delta))}{\epsilon^2}.
\end{equation*}
\end{theorem}
%-----------------------------------------------

\begin{IEEEproof}
Suppose the client is sampled without replacement with probability $r$ at each round. By Lemma~\ref{lemma:gaussian_mechanism} and Lemma~\ref{lemma:rdp_sub}, the $t$-th round of DP$^2$-FedSAM satisfies $(\alpha, \rho_t(\alpha))$-RDP, where 
\begin{equation}
\label{eqn:perround}
    \rho_t(\alpha) = \frac{3.5r^2\alpha}{\sigma^2},
\end{equation}
if $\sigma^2 \geq 0.7$ and $\alpha \leq 1+ (2/3)C^2\sigma^2\log(1/r\alpha(1+ \sigma^2)) $. Then by Lemma~\ref{lemma:rdp_comr}, DP$^2$-FedSAM satisfies $(\alpha, T\rho_t(\alpha))$-RDP after $T$ rounds of training. Next, in order to guarantee $(\epsilon,\delta)$-DP according to Lemma~\ref{lemma:rdp_dp}, we need 
\begin{equation}
\label{eqn:simplify}
    \frac{3.5r^2 T\alpha}{\sigma^2} + \frac{\log(1/\delta)}{\alpha-1} \leq \epsilon.
\end{equation}
Suppose $\alpha$ and $\sigma$ are chosen such that the conditions for \eqref{eqn:perround} are satisfied. Choose $\alpha = 1 + 2\log(1/\delta)/\epsilon$ and rearrange the inequality in \eqref{eqn:simplify}, we need
\begin{equation}
    \sigma^2 \geq \frac{7r^2T(\epsilon + 2\log(1/\delta))}{\epsilon^2}.
\end{equation}
Then using the constraint on $\epsilon$ concludes the proof.
\end{IEEEproof}

\begin{remark}
It is apparent that employing a lower sampling rate $r$ can strengthen privacy protection by diminishing the privacy budget. However, this may lead to a reduction in training performance as fewer clients participate in each communication round. As a result, the choice of $r$ needs to balance these two aspects. 
\end{remark}

%%%%%%%%%%%%%%%%%%%%%%%%%%%%%%%%%%%%
\section{Performance Evaluation}\label{sec:exp_results}
%%%%%%%%%%%%%%%%%%%%%%%%%%%%%%%%%%%%
In this section, we perform extensive experiments to validate the effectiveness of the proposed scheme by using the following common DPFL methods and their fine-tuned analogues as baselines:
\begin{itemize}
    \item DP-FedAvg~\cite{mcmahan2017learning}: The classic variant of FedAvg that achieves client-level DP, where the full local model updates from each client is clipped by a threshold $C$ and then perturbed by adding Gaussian noise from the distribution $\mathcal{N}(0,(C^2\sigma/(rN))\cdot\mathbf{I}_d)$, where $\sigma$ is the noise multiplier and $r$ is the client sampling ratio per round.
    \item DP-FedAvg-FT~\cite{yu2020salvaging}: The fine-tuned version of DP-FedAvg, which locally fine-tunes the aggregated model downloaded from the server.
    \item DP-FedSAM~\cite{shi2023make}: This method uses the SAM optimizer during the local training process and adheres to the same procedures as DP-FedAvg to clip and add noise before uploading the local updates. It has outperformed prior methods and represents the SOTA in DPFL.
    \item DP-FedSAM-FT: The fine-tuned version of DP-FedSAM, which locally fine-tuned the aggregated model downloaded from the server.
    \item CENTAUR~\cite{collins2021exploiting}: This method is proposed for instance-level DP, which focuses on safeguarding the privacy of each instance in any client's dataset. To apply client-level DP to CENTAUR, we modify it to function as a variant of DP$^2$-FedSAM after replacing the SAM optimizer with the standard SGD optimizer.
\end{itemize}

%%%%%%%%%%%%%%%%%%%%%%%%%%%%%%
\subsection{Experimental Setup}
%%%%%%%%%%%%%%%%%%%%%%%%%%%%%%
We evaluate the performance of DP$^2$-FedSAM on two commonly used datasets in DPFL: FEMNIST and CIFAR-10. The FEMNIST dataset is a federated split variant of the EMNIST dataset which comprises 3,550 clients. We randomly select 500 clients to simulate the non-IID data distribution. For CIFAR-10, to simulate the non-IID distribution across clients, we follow \cite{hsu2019measuring} to split the data in a pathological heterogeneous setting characterized by $(N, S)$, where we sample $S$ classes from a total of 10 classes for $N$ clients with disjoint data samples. For both datasets, each client's local data is partitioned into $90\%$ for training and $10\%$ for testing. We use ResNet-18 for FEMNIST and a simple CNN for CIFAR-10 dataset, both pre-trained on ImageNet. The trained models on FEMNIST and CIFAR-10 have 11,181,642 and 667,402 parameters, respectively. The CNN model for CIFAR-10 consists of three 3 × 3 convolutional layers (the first with 64 filters, the second with 128 filters, and the third with 256 filters, each followed by 2 × 2 max pooling and ReLU activation), two fully connected layers (the first with 256 units, the second with 128 units, each followed by ReLU activation), and a final softmax output layer with 10 units for classification.

For FEMNIST and CIFAR-10 datasets, we set the number of communication rounds $T$ to be 200. For all experiments, we set the learning rate for the shared representation extractor $\eta_{h}$ to be 0.1 and for the personal classifier $\eta_{\phi}$ to be 0.005, respectively, decaying at a rate of 0.99 at each communication round. The default momentum is $0.3$ and $0.7$ for FEMNIST and CIFAR-10, respectively. For fair comparisons, we set the total number of local epochs to 2 for DP-FedAvg, DP-FedAvg-FT, DP-FedSAM, and DP-FedSAM-FT. Similarly, for CENTAUR and DP$^2$-FedSAM, we set the local epochs $\tau_h$ and $\tau_{\phi}$ to 2. For CENTAUR and DP$^2$-FedSAM, we choose the last fully connected layer as the head layer. For privacy parameters, we set $\delta=1/N$ by default. Through a unified grid search process from the set $\{1.0, 0.5, 0.1, 0.05, 0.01\}$, a threshold of $C=0.2$ was selected for FEMNIST and $C=0.1$ for CIFAR-10. The perturbation parameter $q$ is set to 0.5 for FEMNIST and 0.1 for CIFAR-10. 

We use the same evaluation metric as prior works~\cite{t2020personalized,zhang2023fedala}, which reports the test accuracy of the best global model for the traditional FL and the average test accuracy of the best local models for personalized FL. For all experiments, we run them three times and report the average and standard deviation of testing accuracies over the final round. In each communication round, the server uniformly samples clients with the client sampling ratio $r=0.05$ to participate in the training process. All algorithms are implemented using Pytorch on an
Ubuntu server with 4 NVIDIA RTX 8000 GPUs.

%%%%%%%%%%%%%%%%%%%%%%%%%%%%%%opocus: 
\subsection{Experimental Results}
%%%%%%%%%%%%%%%%%%%%%%%%%%%%%%
%%%%%%%%%%%%%%%%%%%%%%%%%%%%%%%%%
\begin{figure}[t]
    \subfloat[FEMNIST]{{\includegraphics[width=0.24\textwidth]{ {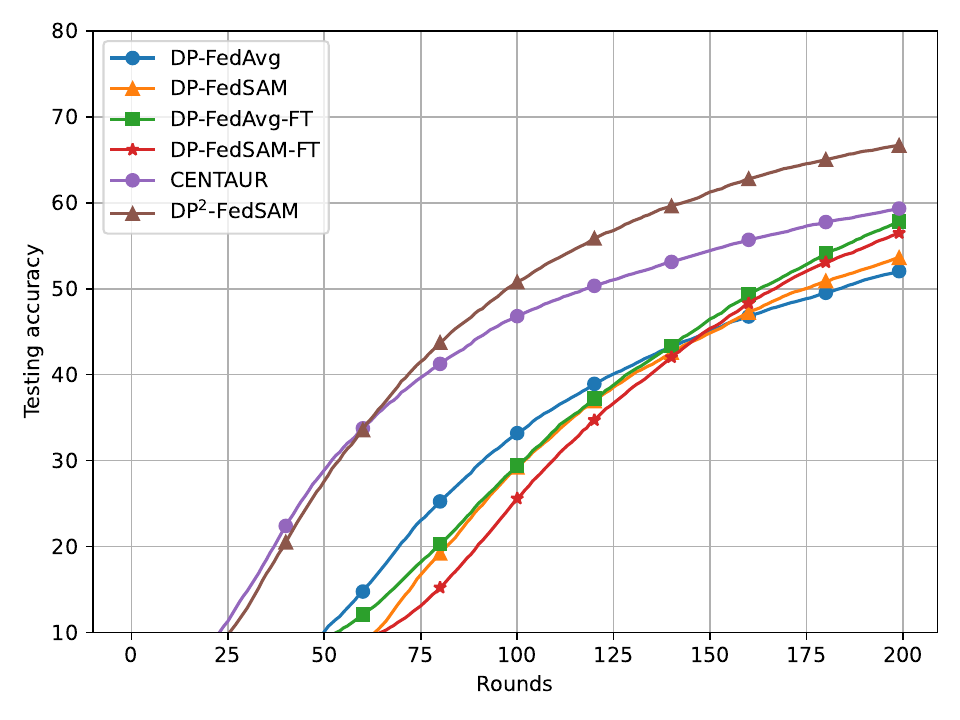}} }\label{fig:fig_femnist_500_5}}
    \subfloat[CIFAR-10 $(N=500, S=5)$]{{\includegraphics[width=0.24\textwidth]{ {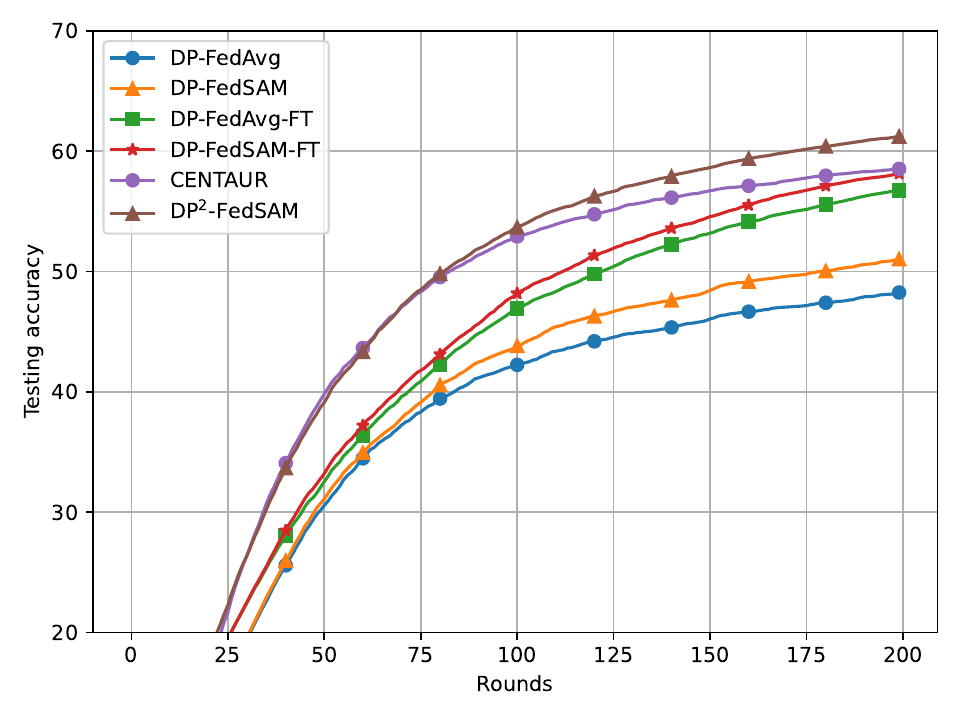}} }\label{fig:fig_500_5}}\\
    \subfloat[CIFAR-10 $(N=1000, S=2)$]{{\includegraphics[width=0.24\textwidth]{ {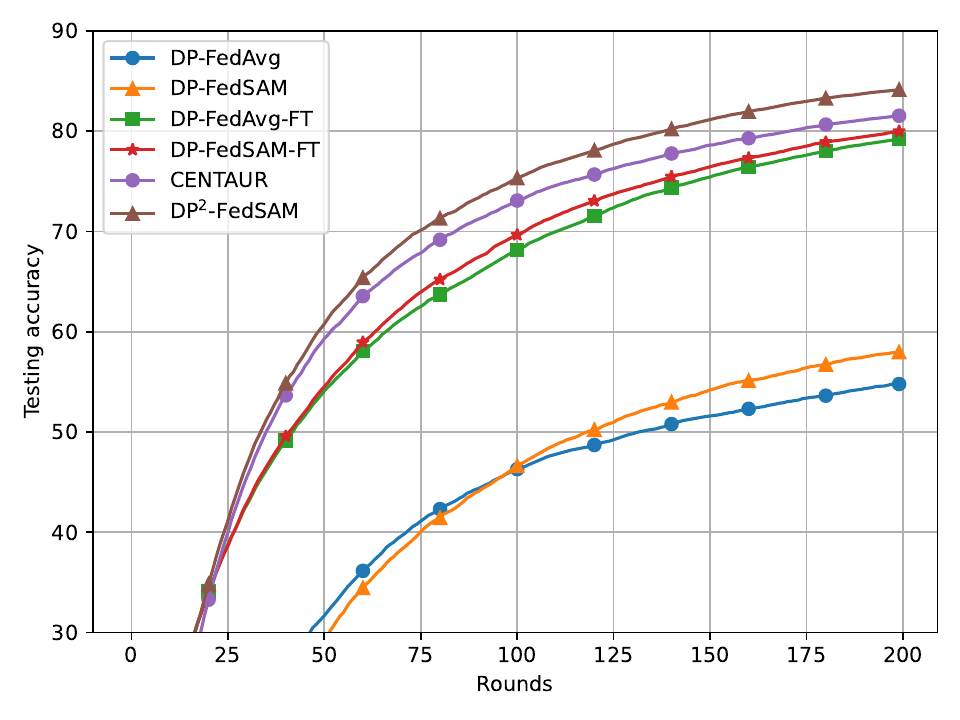}} }\label{fig:fig_1000_2}}
    \subfloat[CIFAR-10 $(N=1000, S=5)$]{{\includegraphics[width=0.24\textwidth]{ {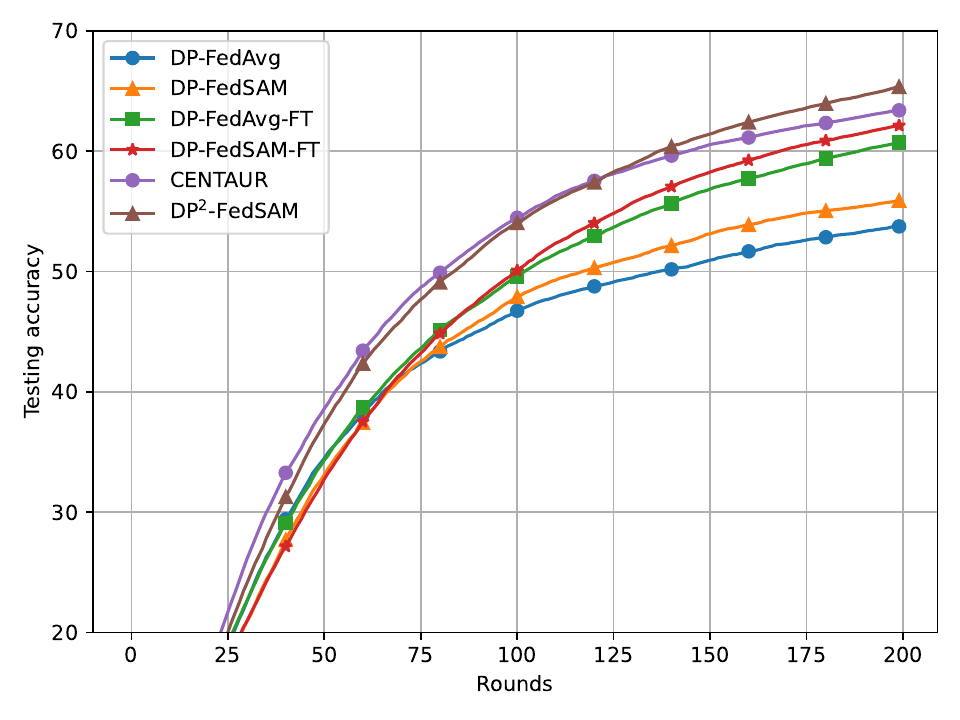}} }\label{fig:fig_1000_5}} 
    \caption{Training performance versus communication round for FEMNIST and CIFAR-10 under $\epsilon= 1.0$.}\label{fig:test_acc_noniid}
\end{figure}
%%%%%%%%%%%%%%%%%%%%%%%%%%%%%%%%%

\subsubsection{Performance comparison under different data distribution settings}
We first compare the testing accuracies of DP$^2$-FedSAM and baselines under various non-IID distribution settings with a fixed privacy budget of $\epsilon=1$. Fig.~\ref{fig:test_acc_noniid} illustrates the testing accuracies over communication rounds, while Table~\ref{tab_hete} summarizes the final average accuracy and standard deviation after $T$ rounds for all schemes. For CIFAR-10 dataset, the results are further segmented based on different combinations of the total number of clients $N$ and the number of classes each client possesses $S$, i.e., $(500, 5)$, $(1000, 2)$, and $(1000, 5)$. 

From the Fig.~\ref{fig:test_acc_noniid}, we have the following observations. First, personalized methods outperform traditional FL methods in heterogeneous data distribution settings. Specifically, DP-FedAvg and DP-FedSAM exhibit the lowest accuracy because they use a shared global model for all clients, which fails to address the individual needs of clients in heterogeneous data settings. Second, fine-tuned variants such as DP-FedAvg-FT and DP-FedSAM-FT show improved performance over their non-fine-tuned counterparts. This improvement is attributed to the additional local fine-tuning, which helps the global model better adapt to individual client data distributions. Third, partial model sharing methods, such as CENTAUR and DP$^2$-FedSAM, outperform fine-tuned methods like DP-FedAvg-FT and DP-FedSAM-FT. This superiority arises because partial model sharing enables a more consistent shared part, mitigating the effect of clipping. Finally, the SAM optimizer further improves the performance by providing flat minima, which helps to mitigate the effects of both clipping and adding noise. For example, DP-FedSAM, DP-FedSAM-FT, and DP$^2$-FedSAM perform better than DP-FedAvg, DP-FedAvg-FT, and CENTAUR.

As summarized in Table~\ref{tab_hete}, DP$^2$-FedSAM outperforms other baselines across different heterogeneous settings. Specifically, DP$^2$-FedSAM achieves an approximate 12\% to 30\% and 14\% enhancement in accuracy over DP-FedAvg for CIFAR-10 and FEMNIST datasets, respectively. Our method consistently shows better performance compared to SOTA methods. For instance, it approximately improves the averaged testing accuracy by around 5\% compared to DP-FedSAM-FT and by about 3\% compared to CENTAUR.

%%%%%%%%%%%%%%%%%%%%%%%%%%%%%%%%%%%%%
\begin{table}[tb]
\caption{Testing accuracy (\%) comparison under different data distribution settings with $\epsilon=1.0$. $N$ represents the total number of clients, and $S$ is the number of classes each client has.}\label{tab_hete}   
    \centering
    \begin{tabular}{@{}lcccc@{}}
        \toprule
            Dataset & FEMNIST & \multicolumn{3}{c}{CIFAR-10} \\
        \midrule
            $(N, S)$  & Non-IID & $(500, 5)$ & $(1000, 2)$ & $(1000, 5)$ \\ 
        \midrule
        DP-FedAvg &  52.2\textsubscript{$\pm$0.7} & 48.3\textsubscript{$\pm$0.8} & 54.8\textsubscript{$\pm$0.5}  & 53.7\textsubscript{$\pm$0.4} \\
        DP-FedAvg-FT &  57.8\textsubscript{$\pm$0.7} & 56.7\textsubscript{$\pm$0.6} & 79.2\textsubscript{$\pm$0.4} & 60.7\textsubscript{$\pm$0.1}\\
        DP-FedSAM &  53.6\textsubscript{$\pm$0.8} & 51.0\textsubscript{$\pm$0.7} & 58.0\textsubscript{$\pm$0.4} & 55.9\textsubscript{$\pm$0.1}\\
        DP-FedSAM-FT &  56.5\textsubscript{$\pm$0.6} & 58.1\textsubscript{$\pm$0.2} & 80.0\textsubscript{$\pm$0.5}& 62.1\textsubscript{$\pm$0.1}\\
        CENTAUR & 59.3\textsubscript{$\pm$0.4} & 58.5\textsubscript{$\pm$0.4} & 81.5\textsubscript{$\pm$0.6}& $63.4\textsubscript{$\pm$0.3}$\\
        DP$^2$-FedSAM & \bf{66.7}\textsubscript{$\pm$0.7} & \bf{61.2}\textsubscript{$\pm$0.2} & \bf{84.1}\textsubscript{$\pm$0.6} & \bf{65.3}\textsubscript{$\pm$0.5}\\
        \bottomrule
    \end{tabular}
\end{table}

\subsubsection{Performance comparison under different privacy budgets}
% \noindent {\bf .}
\begin{table}[t]
\caption{Testing accuracy (\%) comparison under different privacy budgets. A smaller $\epsilon$ indicates a stronger privacy guarantee.}
    \label{tab_priv}
    \centering
    % \resizebox{0.7\linewidth}{!}{
    \begin{tabular}{@{}l@{\extracolsep{5pt}}cccc@{}}
        \toprule
            Dataset & \multicolumn{2}{c}{FEMNIST} & \multicolumn{2}{c}{CIFAR-10}  \\
        \midrule
            $\epsilon$  &  1.0 &  2.0 &  1.0  &  2.0   \\ 
        \midrule
            DP-FedAvg & 52.2\textsubscript{$\pm$0.7} & 62.9\textsubscript{$\pm$1.7} &  54.8\textsubscript{$\pm$0.5} & 65.9\textsubscript{$\pm$0.6}  \\
            DP-FedAvg-FT & 57.8\textsubscript{$\pm$0.7} & 66.7\textsubscript{$\pm$1.2} &  79.2\textsubscript{$\pm$0.4} & 82.8\textsubscript{$\pm$0.1}  \\
            DP-FedSAM & 53.6\textsubscript{$\pm$0.8} & 64.3\textsubscript{$\pm$1.2} &  58.0\textsubscript{$\pm$0.4} & 67.2\textsubscript{$\pm$0.3} \\
            DP-FedSAM-FT & 56.5\textsubscript{$\pm$0.6} & 67.4\textsubscript{$\pm$0.4} &  80.0\textsubscript{$\pm$0.5} & 83.3\textsubscript{$\pm$0.2}  \\
            CENTAUR & 59.3\textsubscript{$\pm$0.4} & 69.6\textsubscript{$\pm$0.7} & 81.5\textsubscript{$\pm$0.6} & 83.1\textsubscript{$\pm$0.5}  \\
            DP$^2$-FedSAM & {\bf66.7}\textsubscript{$\pm$0.7} & {\bf72.2}\textsubscript{$\pm$0.4} & \bf{84.1}\textsubscript{$\pm$0.6} & \bf{85.3}\textsubscript{$\pm$0.8} \\
        \bottomrule
    \end{tabular}
    % }
\end{table}

Table~\ref{tab_priv} provides a comprehensive comparison of testing accuracy for different schemes under varying levels of privacy budget, as indicated by different values of $\epsilon$. For CIFAR-10, we conduct experiments under $(N, S) = (1000, 2)$. A lower value of $\epsilon$ corresponds to a stronger privacy guarantee. The results demonstrate a clear trend where the DP$^2$-FedSAM method consistently outperforms other methods across both FEMNIST and CIFAR-10 datasets for all privacy budget settings. Notably, for the most restrictive privacy setting ($\epsilon = 1.0$), DP$^2$-FedSAM achieves the highest testing accuracies of 66.7\% on FEMNIST and 84.1\% on CIFAR-10, respectively. This trend persists across different values of $\epsilon$, suggesting the robustness of DP$^2$-FedSAM in maintaining high accuracy under stringent privacy constraints.

Other approaches also outperform DP-FedAvg but lag behind DP$^2$-FedSAM, especially under the small privacy budget regime. This indicates that while methods like DP-FedAvg-FT, DP-FedSAM, and DP-FedSAM-FT can enhance accuracy to some extent, they do not achieve the same level of privacy-utility tradeoff as DP$^2$-FedSAM. These results highlight the superiority of DP$^2$-FedSAM in achieving a better privacy-utility tradeoff in DPFL, making it a promising approach for privacy-preserving machine learning applications.

\subsubsection{Effect of low local update norms and consistent local updates.}
%%%%%%%%
\begin{figure}[t]
  \centering
  \includegraphics[width=0.38\textwidth]{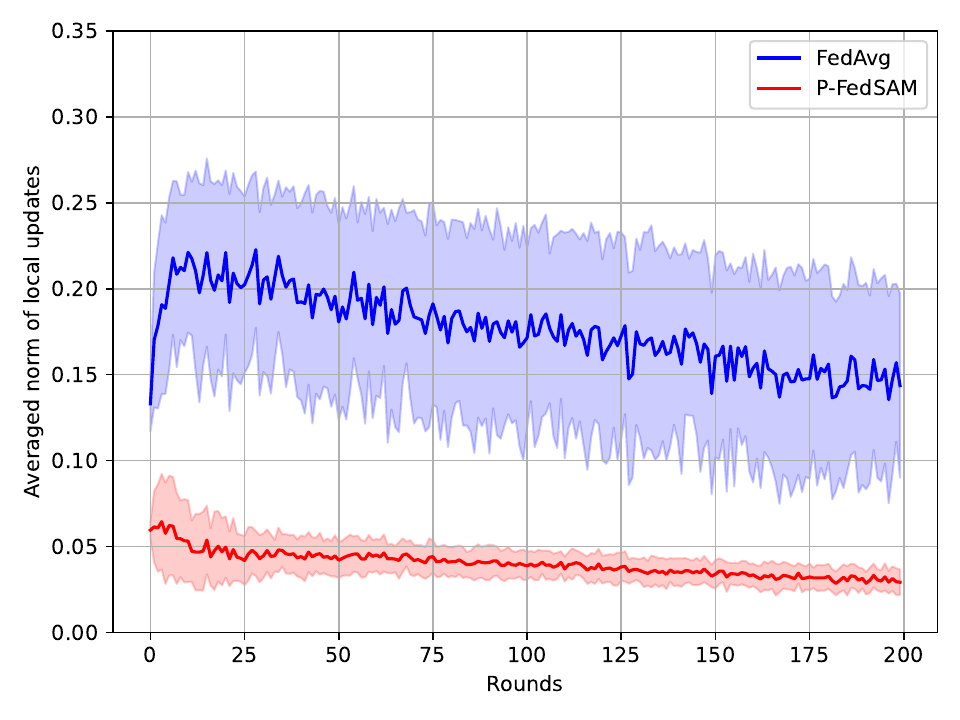}
   \caption{The averaged norm of local updates $\Delta_i^t$ versus communication round.}
   \label{fig:norm} 
\end{figure}
%%%%%%%%
We conduct additional experiments to verify the effectiveness of partial model personalization and SAM. The experiments are conducted on CIFAR-10 with $(N, S)=(1000, 2)$. P-FedSAM is essentially DP$^2$-FedSAM without the mechanisms of clipping and noise addition. In Fig.~\ref{fig:norm}, the solid lines represent the averaged norm of local updates, while the shaded area around the solid line indicates the standard deviation. As shown in  Fig.~\ref{fig:norm}, we have the following observations. First, P-FedSAM significantly reduces the norm of local updates. It verifies that partial model-sharing and SAM can produce low local norms. This inherent characteristic lowers the probability of gradients exceeding the clipping threshold, thereby decreasing the clipping error and promoting more efficient convergence in DP training. Second, the standard deviation/variance of local updates P-FedSAM is much smaller than that in FedAvg. This is due to more consistent partial model-sharing parts and the global flatter minima. The flatter minima demonstrate greater
resilience compared to sharp minima under the same
noise magnitude in DP training. Therefore, these more consistent local updates further mitigate the performance degradation in the heterogeneous data distribution setting.

%%%%%%%%%%%%%%%%%%%%%%%%%%%%%%
\subsection{Ablation Study}
%%%%%%%%%%%%%%%%%%%%%%%%%%%%%%

\begin{figure}[t]
  \centering
  \includegraphics[width=0.38\textwidth]{ {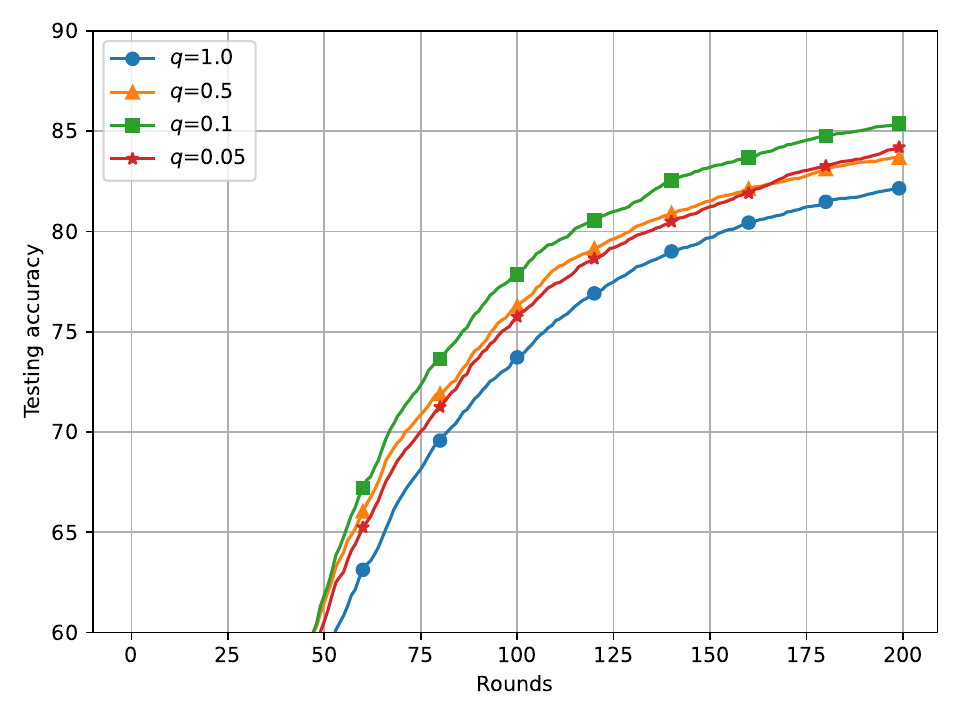}} 
   \caption{Testing accuracy versus communication round on CIFAR-10 dataset with $(N, S)=(1000, 2)$ under different perturbation radius $q$.}
  \label{fig:hyper_q}
\end{figure}

\begin{figure}[t]
  \centering
  \includegraphics[width=0.38\textwidth]{ {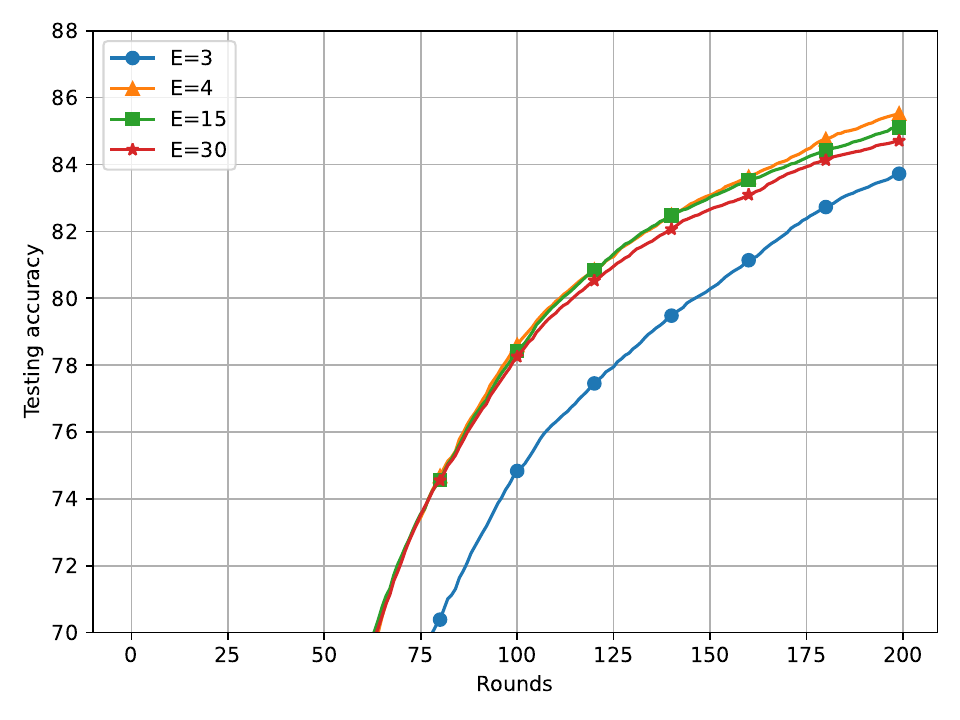}} 
   \caption{Testing accuracy versus communication round on CIFAR-10 dataset with $(N, S)=(1000, 2)$ under different local iterations $\tau_{\phi}$.}
 \label{fig:hyper_E}
\end{figure}

In the following, we illustrate the impact of hyper-parameters in DP$^2$-FedSAM on CIFAR-10 with $(N, S)=(1000, 2)$ in Fig.~\ref{fig:hyper_q} and Fig.~\ref{fig:hyper_E}.

%%%%%%%%
%%%%%%%%

\subsubsection{Impact of perturbation parameter $q$}
The choice of the perturbation radius $q$ significantly influences the performance in DP$^2$-FedSAM. A large $q$ might result in suboptimal performance due to excessive perturbation, while a small $q$ might have little to no impact on the training model. This is because a large perturbation can introduce too much noise, distorting the gradients and leading to poor convergence. Conversely, a small perturbation radius may be insufficient to induce the robustness benefits that perturbation aims to achieve, resulting in minimal impact on the model's performance.

To select an appropriate $q$, we conducted experiments with various perturbation radii from $\{0.05, 0.1, 0.5, 1.0\}$ as illustrated in Fig.~\ref{fig:hyper_q}. The results show that $q = 0.1$ achieves the best performance among the different values tested. This indicates that a moderate level of perturbation is beneficial, providing a balance between introducing necessary robustness and maintaining the integrity of the model updates. 

%%%%%%%%
%%%%%%%%

\subsubsection{Impact of local iteration steps}
We then evaluate the impact of local iteration steps $\tau_{\phi}$ with a fixed $\tau_{h} = 2$. The results in Fig.~\ref{fig:hyper_E} demonstrate that DP$^2$-FedSAM initially improves with an increase in $\tau_{\phi}$, but then performance declines with further increases. This occurs because more local updates can lead to higher client drift, where local models diverge, resulting in severe performance degradation as noted in previous studies \cite{karimireddy2020scaffold}. Client drift is particularly problematic in federated learning with non-IID data distributions, where local data varies significantly between clients. More local updates may cause overfitting to local data, increasing disparity among client models when aggregated, thus harming global model performance. To mitigate this, we experimented with different $\tau_{\phi}$ values to balance minimizing client drift and ensuring sufficient local training. The results indicate that initial increases in $\tau_{\phi}$ improve performance, but beyond a certain point, further increases cause degradation. Hence, we chose 4 epochs, balancing effective local training with maintaining global model coherence, ensuring robust performance across heterogeneous data distributions.

%-------------------------------------------------
\section{Related Work}\label{sec:related}
%-------------------------------------------------

\subsection{Client-level DPFL}

Mcmahan et al.~\cite{McMahan2018learning} first propose DP-FedAvg to ensure client-level DP guarantee by employing the Gaussian mechanism. Following this, Kairouz et al.~\cite{kairouz2021distributed} and Andrew et al.~\cite{andrew2021differentially} achieve client-level DP by discretizing the data and introducing discrete Gaussian noise before conducting secure aggregation. Additionally, they present a novel privacy analysis for the sums of discrete Gaussians. However, the model's utility is unavoidably affected due to clipping and additive noise perturbation. Meanwhile, Zhu et al.~\cite{zhu2022voting} present a voting-based mechanism among the data labels returned from each local model, instead of averaging the gradients. Nonetheless, the AE-DPFL relies on the availability of unlabeled data from the global distribution at the server, a condition that can be difficult to meet in real-world applications. Zhang et al.~\cite{zhang2023commun} propose a novel private federated edge learning with sparsification to provide client-level DP guarantee with intrinsic channel noise while reducing communication and energy overhead and improving model accuracy in wireless FL. Hu et al.~\cite{hu2022federated} integrate the local update sparsification technique into DP-FedAvg and propose a new DPFL scheme that requires a smaller amount of added random noise to achieve the same level of DP. In addition to the local update sparsification technique, Cheng et al.~\cite{cheng2022differentially} leverage bounded local update regularization to further restrict the norm of local updates and reduce the added noise. However, all the aforementioned works~\cite{McMahan2018learning,kairouz2021distributed,andrew2021differentially,hu2022federated,cheng2022differentially} still suffer from model performance degradation due to the inconsistency issue of model updates across clients under data heterogeneity.  

\subsection{Sharpness-Aware Minimization in FL} 

Several recent works have proposed incorporating sharpness-aware minimization (SAM) for better generalization in FL~\cite{caldarola2022improving,qu2022generalized,dai2023fedgamma,sun2023dynamic,fanlocally}. Specifically, Caldarola et al.~\cite{caldarola2022improving} integrate the technique into FL and propose the FedSAM algorithm, which aims to enhance the global model's generalization capabilities and improve overall training performance. Building on this, Qu et al.~\cite{qu2022generalized} introduce a momentum-based variant called MoFedSAM to further refine the approach. Dai et al.~\cite{dai2023fedgamma} introduce a variant of FedSAM named FedGAMMA, inspired by the Scaffold~\cite{karimireddy2020scaffold} framework. Sun et al.~\cite{sun2023dynamic} propose FedSMOO, which adopts a dynamic regularize to guarantee the local optima towards the global objective. At the same time, it employs the global SAM optimizer to search for consistent flat minima. Fan et al.~\cite{fanlocally} propose FedLESAM, an efficient algorithm that locally estimates global perturbations for SAM, optimizing global sharpness while reducing local computational costs. Despite these significant advancements, these methods do not consider rigorous privacy protection for clients nor the incorporation of personalization techniques.

The works that are most related to ours are ~\cite{bietti2022personalization,yang2024dynamic,shi2023make}, and they have studied the personalization strategy and the SAM optimizer in DPFL, respectively. While these methods have their advantages, they differ fundamentally from the strategy we propose. Specifically, DP$^2$-FedSAM distinguishes itself through the following key aspects. 1) Novel integration of partial model personalization and SAM to significantly improve the privacy-utility trade-off in DPFL: Unlike the full model personalization approaches adopted in \cite{bietti2022personalization,yang2024dynamic}, our work uses partial model personalization that shares much fewer numbers of model parameters in each FL round, leading to much less information leakage and higher model accuracy. Compared with \cite{shi2023make} that applies SAM optimizer to the update of the full model in each round, we only apply SAM optimizer to the update of the shared partial model. This selective application aims to not only enhance accuracy but also reduce the additional computational burden typically associated with SAM. Therefore, our method is not a simple combination of two existing strategies. While the personalization strategy and SAM optimizer have been proposed separately in non-privacy settings, integrating and adapting them in a novel way to the DPFL domain is a major technical contribution of our work. 2) Advanced Theoretical Framework: Unlike~\cite{bietti2022personalization} that is limited to a special case of full model personalization with additive model and convex loss, the convergence analysis of our method is much more general and applicable to any partial model penalization strategy (including full model personalization as a special case) and general convex/non-convex loss. Our convergence analysis is also much more challenging than the theoretical analysis of the non-personalized DP-SAM~\cite{shi2023make} approach and significantly improves \cite{yang2024dynamic} that has no convergence guarantee. 3) Empirical Comparison: Through extensive evaluation, DP$^2$-FedSAM has demonstrated a significant improvement in privacy-accuracy trade-off across various settings compared with the SOTA baselines.

%%%%%%%%%%%%%%
\section{Conclusion}\label{sec:conclusion}
%%%%%%%%%%%%%%

In this paper, we have developed DP$^2$-FedSAM, a new DPFL scheme that integrates partial model personalization and sharpness-aware minimization, to enhance accuracy under date heterogeneity. We have provided rigorous analysis on the convergence property and DP guarantee of DP$^2$-FedSAM. Extensive experiments have demonstrated the effectiveness of DP$^2$-FedSAM in balancing privacy and utility in FL, outperforming previous methods. In the future, we plan to conduct more experiments on foundation models and various tasks.

\bibliographystyle{IEEEtran}

\bibliography{main.bib}

% if have a single appendix:
%\appendix[Proof of the Zonklar Equations]
% or
%\appendix  % for no appendix heading
% do not use \section anymore after \appendix, only \section*
% is possibly needed

% use appendices with more than one appendix
% then use \section to start each appendix
% you must declare a \section before using any
% \subsection or using \label (\appendices by itself
% starts a section numbered zero.)
%

% \appendices
% \section{Proof of Theorem~\ref{theorem_conv}\label{appen_proof_conv}}

% % you can choose not to have a title for an appendix
% % if you want by leaving the argument blank
% \section{}
% Appendix two text goes here.

%The authors would like to thank...

% Can use something like this to put references on a page
% by themselves when using endfloat and the captionsoff option.
\ifCLASSOPTIONcaptionsoff
  \newpage
\fi

% trigger a \newpage just before the given reference
% number - used to balance the columns on the last page
% adjust value as needed - may need to be readjusted if
% the document is modified later
%\IEEEtriggeratref{8}
% The "triggered" command can be changed if desired:
%\IEEEtriggercmd{\enlargethispage{-5in}}

% references section

% can use a bibliography generated by BibTeX as a .bbl file
% BibTeX documentation can be easily obtained at:
% http://mirror.ctan.org/biblio/bibtex/contrib/doc/
% The IEEEtran BibTeX style support page is at:
% http://www.michaelshell.org/tex/ieeetran/bibtex/
%\bibliographystyle{IEEEtran}
% argument is your BibTeX string definitions and bibliography database(s)
%\bibliography{IEEEabrv,../bib/paper}
%
% <OR> manually copy in the resultant .bbl file
% set second argument of \begin to the number of references
% (used to reserve space for the reference number labels box)
% \bibliographystyle{IEEEtran}
% \bibliography{main.bib}

\newpage
\onecolumn

% The paper headers
% \markboth{}%
% {Shell \MakeLowercase{\textit{et al.}}: }

% \IEEEtitleabstractindextext{%
% }
% % \onecolumn
% % make the title area
% \maketitle

\appendices
%%%%%%%%%%%%%%%%%%%%%%%%%%%%%%%%%%%%%%%%%%%%%%%%%%%%%%%%%%%%%%%%%%%%%%%%%%%%%%%%%%%%%%%%%%%%%%%%%%%%%%%%%%%%%%%%
%%%%%%%%%%%
\section{Convergence analysis of DP$^2$-FedSAM}\label{appen:proof_conv}
%%%%%%%%%%%
%%%%%%%%%%%%%%%%%%%%%%%%%%%
\subsection{Notions}
%%%%%%%%%%%%%%%%%%%%%%%%%%%
For ease of notion, let $\check{\phi}_{i}^{t,s}$, $\check{h}_i^{t,s}$ denote the virtual sequences as the SAM/SGD updates following Algorithm~\ref{alg_dp2fedsam}, regardless of whether they are selected. Thus, for the selected client $i\in\mathcal{S}^t$, we have $h_{i}^{t,s} = \check{h}_{i}^{t,s}$ and $\phi_{i}^{t,s} = \check{\phi}_{i}^{t,s}$. Note that the random variables $\bar{\phi}_{i}^{t,s}$, $\bar{h}_i^{t,s}$ are independent of the random selection $\mathcal{S}^t$. Then, we have the following update roles for selected clients $i\in\mathcal{S}^t$ in Algorithm~\ref{alg_dp2fedsam} as follows
\begin{gather*}
    h_i^{t+1} = h_i^t - \eta_h\sum_{s=0}^{\tau_h-1}\widetilde{\nabla}_{h}F_i( {\phi}_{i}^{t}, \check{h}_{i}^{t,s}),\\
    \phi_i^{t+1} = \phi_i^t - {\eta_{\phi}}\sum_{s=0}^{\tau_{\phi}-1}\widetilde{\nabla}_{\phi}F_i( \check{\phi}_{i}^{t,s}+p(\check{\phi}_{i}^{t,s}), \check{h}_{i}^{t+1}).
\end{gather*}
where $p(\check{\phi}_{i}^{t,s})$ is given by
\begin{equation*}
      p( \check{\phi}_{i}^{t,s}) = q \frac{\widetilde{\nabla}_{ {\phi}}F_i( \check{\phi}_{i}^{t,s}, \check{h}_{i}^{t+1})}{\norm{\widetilde{\nabla}_{{\phi}}F_i( \check{\phi}_{i}^{t,s}, \check{h}_{i}^{t+1})}_2}.
\end{equation*}
We use $z_i^t$ to denote the Gaussian noise $\mathcal{N}(0,\frac{C^2\sigma^2\mathbf{I}_d}{r})$. Then we have
\begin{align*}
    \Delta_{i}^{t}&= {\phi}_i^{t,\tau} - {\phi}^{t} = -{\eta_{\phi}}\sum_{s=0}^{\tau_{\phi}-1}\widetilde{\nabla}_{\phi}F_i( \check{\phi}_{i}^{t,s}+p(\check{\phi}_{i}^{t,s}), \check{h}_{i}^{t+1}),\\
    \hat{\Delta}_{i}^{t}& = \Delta_{i}^{t}\cdot\min\left(1,\frac{C}{\norm{\Delta_{i}^{t}}_2}\right)+z_i^t.
\end{align*}
The server update rule is given by
\begin{equation*}
    {\phi}^{t+1} =  {\phi}^{t} + \frac{1}{rN}\sum_{i\in\mathcal{S}^{t}} \hat{\Delta}_{i}^{t}.
\end{equation*}
We use the notation $\check{\Delta}_{\phi}^{t}$ as the analogue of ${\Delta}_{\phi}^{t}$ with the virtual variable $\check{H}^{t+1}$ and define the following notions for convenience: 
\begin{align*}
    \tilde{\Delta}_{i}^{t}& = -{\eta_{\phi}}\sum_{s=0}^{\tau_{\phi}-1}\widetilde{\nabla}_{\phi}F_i( \check{\phi}_{i}^{t,s}+p(\check{\phi}_{i}^{t,s}), \check{h}_{i}^{t+1})\cdot\alpha_i^t,\\
     \bar{\Delta}_{i}^{t}& = -{\eta_{\phi}}\sum_{s=0}^{\tau_{\phi}-1}\widetilde{\nabla}_{\phi}F_i( \check{\phi}_{i}^{t,s}+p(\check{\phi}_{i}^{t,s}), \check{h}_{i}^{t+1})\cdot\bar{\alpha}^t,\\
     \dot{\Delta}_{i}^{t}& = -{\eta_{\phi}}\sum_{s=0}^{\tau_{\phi}-1}{\nabla}_{\phi}F_i( \check{\phi}_{i}^{t,s}+p(\check{\phi}_{i}^{t,s}), \check{h}_{i}^{t+1})\cdot\bar{\alpha}^t,
\end{align*}
where
\begin{align*}
    \alpha_i^t& = \min(1,\frac{C}{\eta_{\phi}\norm{\sum_{s=0}^{\tau_{\phi}-1}\widetilde{\nabla}_{\phi}F_i( \check{\phi}_{i}^{t,s}+p(\check{\phi}_{i}^{t,s}), \check{h}_{i}^{t+1})}})\\
     \bar{\alpha}^t& = \frac{1}{N}\sum_{i=1}^{N}\alpha_i^t,\quad
     \tilde{\alpha}^t = \frac{1}{N}\sum_{i=1}^{N}|\alpha_i^t-\bar{\alpha}^t|.
\end{align*}
%%%%%%%%%%%%%%%%%%%%%%%%%%%%%
\subsection{Useful Lemmas}
%%%%%%%%%%%%%%%%%%%%%%%%%%%%%

\setcounter{lemma}{5}
%%%%%%%%%%%%%%%%%%%%%%%%%%%%%%%%%%%%%
\begin{lemma}[Cauchy-Schwarz inequality]\label{lemma_cau_sch_in}
For arbitrary set of $n$ vectors $\{\bm{a}_i\}_{i=1}^n$, $\bm{a}_i \in \mathbb{R}^d$,
\begin{equation*}
    \norm{\sum_{i=1}^n\bm{a}_i}^2\leq n\sum_{i=1}^n\norm{\bm{a}_i}^2
\end{equation*}
\end{lemma}
%%%%%%%%%%%%%%%%%%%%%%%%%%%%%%%%%%%%

%%%%%%%%%%%%%%%%%%%%%%%%%%%%%%%%
\begin{lemma}\label{lemma_in_inn_prod}
For given two vectors $\bm{a},\bm{b}\in \mathbb{R}^d$,
\begin{equation*}
    2\langle \bm{a},\bm{b}\rangle \leq \gamma\norm{\bm{a}}^2 +\gamma^{-1}\norm{\bm{b}}^2, \forall \gamma \geq 0.
\end{equation*}
\end{lemma}
%%%%%%%%%%%%%%%%%%%%%%%%%%%%%%%
%%%%%%%%%%%%%%%%%%%%%%%%%%%%%%%
\begin{lemma}[Bounded $\mathcal{T}_{1,\phi}$]\label{lemma_t_1_phi} For $\mathcal{T}_{1,\phi}$, we have,

\begin{align*}
    \mathbb{E}_t[\mathcal{T}_{1,\phi}]  & \leq \eta_{\phi}\tau_{\phi}\tilde{\alpha}^tL_{\phi}^2q^2 -\frac{\bar{\alpha}^t\eta_{\phi}\tau_{\phi}}{2}\norm{\nabla_{\phi}F( {\phi}^{t}, \check{H}^{t+1})}^2- \frac{\eta_{\phi}\bar{\alpha}^t}{2\tau_{\phi}}\norm{\frac{1}{\eta_{\phi}\bar{\alpha}^tN}\sum_{i=1}^{N}\dot{\Delta}_i^t}^2\\
    &\quad+{L_{\phi}^2\eta_{\phi}\tau_{\phi}\bar{\alpha}^t}q^2+36\tau_{\phi}\eta_{\phi}^2(2\sigma_{\phi}^2+G^2)+L_{\phi}^2\eta_{\phi}\tau_{\phi}\bar{\alpha}^t18\eta_{\phi}^2\tau_{\phi}^2\Big(\sigma_{\phi}^2+\delta^2+\|{\nabla}_{\phi}F(\phi^{t},H^{t+1})\|^2\Big).
\end{align*}
\end{lemma}
%%%%%%%%%%%%%%%%%%%%%%%%%%%%%%%%%%%%
\begin{IEEEproof}
For client $i\in\mathcal{S}^t$, we have $\check{\phi}_{i}^{t,s}={\phi}_{i}^{t,s}$. Thus, we have
\begin{align}
    \mathbb{E}_t[\mathcal{T}_{1,\phi}]& = \mathbb{E}_t\langle \nabla_{\phi}F( {\phi}^{t}, \check{H}^{t+1}),  {\phi}^{t+1}- {\phi}^{t}\rangle\nonumber\\
    & = \mathbb{E}_t\langle \nabla_{\phi}F( {\phi}^{t}, \check{H}^{t+1}),  \frac{1}{rN}\sum_{i\in\mathcal{S}^t}(\tilde{\Delta}_i^t+z_i^t)\rangle \nonumber\\
    & = \langle \nabla_{\phi}F( {\phi}^{t}, \check{H}^{t+1}),  \mathbb{E}_t\frac{1}{rN}\sum_{i\in\mathcal{S}^t}\tilde{\Delta}_i^t\rangle \nonumber\\
    & = \langle \nabla_{\phi}F( {\phi}^{t}, \check{H}^{t+1}),  \mathbb{E}_t\frac{1}{N}\sum_{i=1}^{N}\tilde{\Delta}_i^t-\bar{\Delta}_i^t\rangle + \langle\nabla_{\phi}F( {\phi}^{t}, \check{H}^{t+1}),  \mathbb{E}_t\frac{1}{N}\sum_{i=1}^{N}\bar{\Delta}_i^t\rangle.\label{eq_delta_iner}
\end{align}
For the first term, we obtain

\begin{align*}
    \mathbb{E}_t\langle\nabla_{\phi}F( {\phi}^{t}, \check{H}^{t+1}),  \mathbb{E}_t\frac{1}{N}\sum_{i=1}^{N}\tilde{\Delta}_i^t-\bar{\Delta}_i^t\rangle  &= -\mathbb{E}_t\langle\nabla_{\phi}F( {\phi}^{t}, \check{H}^{t+1}),  \mathbb{E}_t\frac{1}{N}\sum_{i=1}^{N}\sum_{s=0}^{\tau_{\phi}-1}\eta_{\phi}(\alpha_i^t-\bar{\alpha}^t)\widetilde{\nabla}_{\phi}F_i( \check{\phi}_{i}^{t,s}+p(\check{\phi}_{i}^{t,s}), \check{h}_{i}^{t+1})\rangle\nonumber\\
    &= -\frac{\eta_{\phi}\tau_{\phi}}{N}\sum_{i=1}^{N}\mathbb{E}_t(\alpha_i^t-\bar{\alpha}^t)\langle\nabla_{\phi}F_i( {\phi}^{t}, h_i^{t+1}), {\nabla}_{\phi}F_i( \check{\phi}_{i}^{t,s}+p(\check{\phi}_{i}^{t,s}), \check{h}_{i}^{t+1})\rangle\nonumber\\
    &\labelrel={in_iner} \frac{\eta_{\phi}\tau_{\phi}}{N}\sum_{i=1}^{N}\mathbb{E}_t(\alpha_i^t-\bar{\alpha}^t)\Big(\frac{1}{2}\norm{\nabla F_i( {\phi}^{t}, h_i^{t+1})-{\nabla}_{\phi}F_i( \check{\phi}_{i}^{t,s}+p(\check{\phi}_{i}^{t,s}), \check{h}_{i}^{t+1})}^2\notag\\
    & \quad-\frac{1}{2}\big(\norm{\nabla F_i( {\phi}^{t}, h_i^{t+1})}^2+\norm{{\nabla}_{\phi}F_i( \check{\phi}_{i}^{t,s}+p(\check{\phi}_{i}^{t,s}), \check{h}_{i}^{t+1})}^2 \big)\Big)\notag\\
    &\labelrel\leq{in_p_g} \eta_{\phi}\tau_{\phi}\tilde{\alpha}^tL_{\phi}^2q^2,\notag
\end{align*}

where~(\ref{in_iner}) holds due to $-\langle a,b \rangle=-\frac{1}{2}\norm{a}^2-\frac{1}{2}\norm{b}^2+\frac{1}{2}\norm{a-b}^2$,~(\ref{in_p_g}) follows from Assumption \ref{assum_smooth}. For the second term in~\eqref{eq_delta_iner}, we get

\begin{align*}
    \langle\nabla_{\phi}F( {\phi}^{t}, \check{H}^{t+1}),  \mathbb{E}_t\frac{1}{N}\sum_{i=1}^{N}\bar{\Delta}_i^t\rangle &= \langle\nabla_{\phi}F( {\phi}^{t}, \check{H}^{t+1}),  \mathbb{E}_t\frac{1}{N}\sum_{i=1}^{N}\dot{\Delta}_i^t\rangle\\
    &\labelrel\leq{in_a1} \frac{-\bar{\alpha}^t\eta_{\phi}\tau_{\phi}}{2}\norm{\nabla_{\phi}F( {\phi}^{t}, \check{H}^{t+1})}^2\\
    & \quad- \frac{\eta_{\phi}\bar{\alpha}^t}{2\tau_{\phi}}\norm{\frac{1}{\eta_{\phi}\bar{\alpha}^tN}\sum_{i=1}^{N}\dot{\Delta}_i^t}^2  + \frac{\eta_{\phi}\bar{\alpha}^t}{2}\underbrace{\mathbb{E}\norm{\sqrt{\tau_{\phi}}\nabla_{\phi}F( {\phi}^{t}, \check{H}^{t+1}) + \frac{1}{\eta_{\phi}\bar{\alpha}^tN\sqrt{\tau_{\phi}} }\sum_{i=1}^{N}\dot{\Delta}_i^t}^2}_{A_1},
\end{align*}
where~(\ref{in_a1}) follows from $\langle a,b\rangle = -\frac{1}{2}\norm{a}^2-\frac{1}{2}\norm{b}^2+\frac{1}{2}\norm{a+b}^2$. For $A_1$, we have

\begin{align*}
    A_1 & = \tau_{\phi}\mathbb{E}\norm{\nabla_{\phi}F( {\phi}^{t},\check{H}^{t+1}) - \frac{1}{N\tau_{\phi}}\sum_{i=1}^{N}\sum_{s=0}^{\tau_{\phi}-1}{\nabla}_{\phi}F_i( \tilde{\phi}_{i}^{t,s}+p(\tilde{\phi}_{i}^{t,s}), \tilde{h}_{i}^{t+1})}^2\\
    & = \tau_{\phi}\mathbb{E}\norm{\frac{1}{N\tau_{\phi}}\sum_{i=1}^{N}\sum_{s=0}^{\tau_{\phi}-1}\nabla_{\phi}F_i( {\phi}^{t}, \tilde{h}_{i}^{t+1}) - {\nabla}_{\phi}F_i( \tilde{\phi}_{i}^{t,s}+p(\tilde{\phi}_{i}^{t,s}), \tilde{h}_{i}^{t+1})}^2\\
    & \leq \frac{1}{N}\sum_{i=1}^{N}\sum_{s=0}^{\tau_{\phi}-1}\mathbb{E}\norm{\nabla_{\phi}F_i( {\phi}^{t},\tilde{h}_i^{t+1}) - {\nabla}_{\phi}F_i( \tilde{\phi}_{i}^{t,s}+p(\tilde{\phi}_{i}^{t,s}), \tilde{h}_{i}^{t+1})}^2\\
    & = \frac{1}{N}\sum_{i=1}^{N}\sum_{s=0}^{\tau_{\phi}-1}\mathbb{E}\|{\nabla}_{\phi}F_i( \tilde{\phi}_{i}^{t,s}+p(\tilde{\phi}_{i}^{t,s}), \tilde{h}_{i}^{t+1})-{\nabla}_{\phi}F_i( \tilde{\phi}_{i}^{t,s}, \tilde{h}_{i}^{t+1})+{\nabla}_{\phi}F_i( \tilde{\phi}_{i}^{t,s}, \tilde{h}_{i}^{t+1})-\nabla_{\phi}F_i( {\phi}^{t},\tilde{h}_i^{t+1})\|^2\\
    & \labelrel\leq{in_a1_var} \frac{L_{\phi}^2}{N}\sum_{i=1}^{N}\sum_{s=0}^{\tau_{\phi}-1}[2q^2+2\mathbb{E}\norm{\tilde{\phi}_{i}^{t,s}-{\phi}^{t}}^2]\\
    & \labelrel\leq{in_a1_var_bound} {2L_{\phi}^2\tau_{\phi}}\left[q^2+36\tau_{\phi}\eta_{\phi}^2(2\sigma_{\phi}^2+G^2)+18\eta_{\phi}^2\tau_{\phi}^2\Big(\sigma_{\phi}^2+\delta^2+\|{\nabla}_{\phi}F(\phi^{t},H^{t+1})\|^2\Big)\right],
\end{align*}

where~(\ref{in_a1_var}) follows from Assumption~\ref{assum_smooth},~(\ref{in_a1_var_bound}) holds due to Lemma~\ref{lemma_bound_loc_upd}. Thus, we have

\begin{align*}
     \mathbb{E}_t[\mathcal{T}_{1,\phi}]  & \leq \eta_{\phi}\tau_{\phi}\tilde{\alpha}^tL_{\phi}^2q^2 -\frac{\bar{\alpha}^t\eta_{\phi}\tau_{\phi}}{2}\norm{\nabla_{\phi}F( {\phi}^{t}, \check{H}^{t+1})}^2 - \frac{\eta_{\phi}\bar{\alpha}^t}{2\tau_{\phi}}\norm{\frac{1}{\eta_{\phi}\bar{\alpha}^tN}\sum_{i=1}^{N}\dot{\Delta}_i^t}^2\\
    &\quad+{L_{\phi}^2\eta_{\phi}\tau_{\phi}\bar{\alpha}^t}\left[q^2+36\tau_{\phi}\eta_{\phi}^2(2\sigma_{\phi}^2+G^2)+18\eta_{\phi}^2\tau_{\phi}^2\Big(\sigma_{\phi}^2+\delta^2+\|{\nabla}_{\phi}F(\phi^{t},H^{t+1})\|^2\Big)\right].
\end{align*}

\end{IEEEproof}
%------------------------------------------------------
\begin{lemma}[Bounded $\mathcal{T}_{2,\phi}$]\label{lemma_t_2_phi} For $\mathcal{T}_{2,\phi}$, we have,
\begin{equation*}
    \mathbb{E}_t[\mathcal{T}_{2,\phi}] \leq 3L_{{\phi}}\eta_{\phi}^2\tau_{\phi}(\sigma_{\phi}^{2}+L_{\phi}^2p^2+G^2)+ \frac{L_{\phi}\sigma^2C^2d_1^2}{r^2N^2}.
\end{equation*}
\end{lemma}
%---------------------------------------------------------
\begin{IEEEproof} Using $\mathbb{E}\norm{x}^2 = \norm{\mathbb{E}[x]}^2 + \mathbb{E}\norm{x-\mathbb{E}[x]}^2$, we get

\begin{align*}
    \mathbb{E}_t[\mathcal{T}_{2,\phi}]& = L_{{\phi}}\mathbb{E}_t\norm{ {\phi}^{t+1}- {\phi}^{t}}^2 \\
    &= L_{{\phi}}\mathbb{E}_t\norm{ \frac{1}{rN}\sum_{i\in\mathcal{S}^t}(\tilde{\Delta}_i^t+z_i^t)}^2\\
    &= L_{{\phi}}\mathbb{E}_t\norm{ \frac{1}{rN}\sum_{i\in\mathcal{S}^t}\tilde{\Delta}_i^t}^2 + \frac{L_{\phi}\sigma^2C^2d_1^2}{r^2N^2}\\
    &\leq L_{{\phi}}\mathbb{E}_t\norm{ \frac{\eta_{\phi}}{rN}\sum_{i\in\mathcal{S}^t}\sum_{s=0}^{\tau_{\phi}-1}\widetilde{\nabla}_{\phi}F_i( \tilde{\phi}_{i}^{t,s}+p(\tilde{\phi}_{i}^{t,s}), \tilde{h}_{i}^{t+1})\cdot\alpha_i^t}^2 + \frac{L_{\phi}\sigma^2C^2d_1^2}{r^2N^2}\\
    &\leq  \frac{L_{{\phi}}\eta_{\phi}^2}{rN}\sum_{i\in\mathcal{S}^t}\mathbb{E}_t\|[\sum_{s=0}^{\tau_{\phi}-1}\widetilde{\nabla}_{\phi}F_i( \tilde{\phi}_{i}^{t,s}+p(\tilde{\phi}_{i}^{t,s}), \tilde{h}_{i}^{t+1})-{\nabla}_{\phi}F_i( \tilde{\phi}_{i}^{t,s}+p(\tilde{\phi}_{i}^{t,s}), \tilde{h}_{i}^{t+1})\\
    &\quad+{\nabla}_{\phi}F_i( \tilde{\phi}_{i}^{t,s}+p(\tilde{\phi}_{i}^{t,s}), \tilde{h}_{i}^{t+1})-{\nabla}_{\phi}F_i( \tilde{\phi}_{i}^{t,s}, \tilde{h}_{i}^{t+1})+{\nabla}_{\phi}F_i( \tilde{\phi}_{i}^{t,s}, \tilde{h}_{i}^{t+1})]\|^2+ \frac{L_{\phi}\sigma^2C^2d_1^2}{r^2N^2}\\
    &\leq 3L_{{\phi}}\eta_{\phi}^2\tau_{\phi}(\sigma_{\phi}^{2}+L_{\phi}^2q^2+G^2)+ \frac{L_{\phi}\sigma^2C^2d_1^2}{r^2N^2}.
\end{align*}

\end{IEEEproof}

%%%%%%%%%%%%%%%%%%%%%%%%%%%%%%%%%%%%
\begin{lemma}[Bounded $\mathcal{T}_{1,h}$](Claim 9,~\cite{pillutla2022federated})\label{lemma_t_h} Assume that $\eta_h\tau_hL_h\leq1/8$, we have

\begin{equation*}
     \mathbb{E}_t[\mathcal{T}_{1,h}] \leq -\frac{\eta_h\tau_hr}{8n}\mathbb{E}\sum_{i=1}^{N}\norm{\nabla_{h} F_i(\phi^t,h_i^t)}^2 + \frac{\eta_h^2\tau_h^2L_h\sigma_h^2r}{2} + 4\eta_h^3L_h\tau_h^2(\tau_h-1)\sigma_h^2r.
\end{equation*}

\end{lemma}
%%%%%%%%%%%%%%%%%%%%%%%%%%%%%%%%%%%%
%%%%%%%%%%%%%%%%%%%%%%%%%%%%%%%%%%%
\begin{lemma}[Bounded $\mathcal{T}_{2,h}$](Claim 8,~\cite{pillutla2022federated})\label{lemma_t_3_phi} For $\mathcal{T}_{2,h}$, we have
\begin{equation*}
     \mathbb{E}_t[\mathcal{T}_{2,h}] \leq 8\eta_{h}^2\tau_h^2L_h\chi^2(1-r)\frac{1}{n}\sum_{i=1}^{N}\norm{\nabla_{h} F_i(\phi^t,h_i^t)}^2 + 4\chi^2\eta_h^2\tau_h^2L_h\sigma_h^2(1-r).
\end{equation*}
\end{lemma}
%%%%%%%%%%%%%%%%%%%%%%%%%%%%%%%%%%%%

%%%%%%%%%%%%%%%%%%%%%%%%%%%%%%%%%%%%
\begin{lemma}[Bounded local updates]\label{lemma_bound_loc_upd}Under Assumptions~\ref{assum_smooth},~\ref{assum_bound_var},~\ref{assum_bound_grad}, we have
\begin{align*} 
    \frac{1}{N}\sum_{i=1}^{N}\mathbb{E}_t\norm{{\phi}_i^{t,s}-\phi^t}^2 \leq 36\tau_{\phi}\eta_{\phi}^2(2\sigma_{\phi}^2+G^2)+18\eta_{\phi}^2\tau_{\phi}^2\Big(\sigma_{\phi}^2+\delta^2+\|{\nabla}_{\phi}F(\phi^{t},H^{t+1})\|^2\Big).
\end{align*}
\end{lemma}
%%%%%%%%%%%%%%%%%%%%%%%%%%%%%%%%%%%%
\begin{IEEEproof}
According to Lemma~\ref{lemma_in_inn_prod}, we obtain

\begin{align*}
        &\frac{1}{N}\sum_{i=1}^{N}\mathbb{E}_t\norm{{\phi}_i^{t,s}-\phi^t}^2 =  \frac{1}{N}\sum_{i=1}^{N}\mathbb{E}_t\norm{{\phi}_i^{t,s-1}-\eta_{\phi}\widetilde{\nabla}_{\phi}F_i(\phi_{i}^{t,s-1}+p(\phi_{i}^{t,s-1}),h_i^{t+1})-\phi^t}^2\\
        & = \frac{1}{N}\sum_{i=1}^{N}\|{\phi}_i^{t,s-1}-\phi^t-\eta_{\phi}\Big(\widetilde{\nabla}_{\phi}F_i(\phi_{i}^{t,s-1}+p(\phi_{i}^{t,s-1}),h_i^{t+1})-\widetilde{\nabla}_{\phi}F_i(\phi_{i}^{t,s-1},h_i^{t+1})\\
        &\quad + \widetilde{\nabla}_{\phi}F_i(\phi_{i}^{t,s-1},h_i^{t+1})-{\nabla}_{\phi}F_i(\phi_{i}^{t,s-1},h_i^{t+1})+{\nabla}_{\phi}F_i(\phi_{i}^{t,s-1},h_i^{t+1}) - {\nabla}_{\phi}F(\phi^{t},H^{t+1})+{\nabla}_{\phi}F(\phi^{t},H^{t+1})\Big)\|^2\\
        &\leq \mathcal{T}_{2,\phi}^{\prime\prime}+\mathcal{T}_{2,\phi}^{\prime\prime\prime},
\end{align*}
where
\begin{align*}
    \mathcal{T}_{2,\phi}^{\prime\prime} = (1+\frac{1}{2\tau_{\phi}-1})\frac{1}{N}\sum_{i=1}^{N}\|{\phi}_i^{t,s-1}-\phi^t- \eta_{\phi}\Big(\widetilde{\nabla}_{\phi}F_i(\phi_{i}^{t,s-1}+p(\phi_{i}^{t,s-1}),h_i^{t+1})-\widetilde{\nabla}_{\phi}F_i(\phi_{i}^{t,s-1},h_i^{t+1})\Big)\|^2,
\end{align*}
and
\begin{align*}
    \mathcal{T}_{2,\phi}^{\prime\prime\prime} =  \frac{2\tau_{\phi}\eta_{\phi}^2}{N}\sum_{i=1}^{N}\|\widetilde{\nabla}_{\phi}F_i(\phi_{i}^{t,s-1},h_i^{t+1})-{\nabla}_{\phi}F_i(\phi_{i}^{t,s-1},h_i^{t+1})+{\nabla}_{\phi}F_i(\phi_{i}^{t,s-1},h_i^{t+1}) - {\nabla}_{\phi}F(\phi^{t},H^{t+1})+{\nabla}_{\phi}F(\phi^{t},H^{t+1})\Big)\|^2.
\end{align*}
For $\mathcal{T}_{2,\phi}^{\prime\prime}$, we get
\begin{align*}
    \mathcal{T}_{2,\phi}^{\prime\prime}&\leq (1+\frac{1}{2\tau_{\phi}-1})\frac{2}{N}\sum_{i=1}^{N}\Big(\mathbb{E}\|{\phi}_i^{t,s-1}-\phi^t\|^2+\eta_{\phi}^2\norm{\widetilde{\nabla}_{\phi}F_i(\phi_{i}^{t,s-1}+p(\phi_{i}^{t,s-1}),h_i^{t+1})-\widetilde{\nabla}_{\phi}F_i(\phi_{i}^{t,s-1},h_i^{t+1}}^2)\\
    &= (1+\frac{1}{2\tau_{\phi}-1})\frac{2}{N}\sum_{i=1}^{N}\Big(\mathbb{E}\|{\phi}_i^{t,s-1}-\phi^t\|^2+\eta_{\phi}^2\|\widetilde{\nabla}_{\phi}F_i(\phi_{i}^{t,s-1}+p(\phi_{i}^{t,s-1}),h_i^{t+1})-{\nabla}_{\phi}F_i(\phi_{i}^{t,s-1}+p(\phi_{i}^{t,s-1}),h_i^{t+1})\\
    &\quad+{\nabla}_{\phi}F_i(\phi_{i}^{t,s-1}+p(\phi_{i}^{t,s-1}),h_i^{t+1})-\widetilde{\nabla}_{\phi}F_i(\phi_{i}^{t,s-1},h_i^{t+1}) +{\nabla}_{\phi}F_i(\phi_{i}^{t,s-1},h_i^{t+1})\|^2\Big)\\
    &\leq (1+\frac{1}{2\tau_{\phi}-1})\frac{2}{N}\sum_{i=1}^{N}\Big(\mathbb{E}\|{\phi}_i^{t,s-1}-\phi^t\|^2 + 3\eta_{\phi}^2(2\sigma_{\phi}^2+G^2)\Big).
\end{align*}
For $\mathcal{T}_{2,\phi}^{\prime\prime\prime}$, we have
\begin{align*}
    \mathcal{T}_{2,\phi}^{\prime\prime\prime}\leq {6\tau_{\phi}\eta_{\phi}^2}\Big(\sigma_{\phi}^2+\delta^2+\|{\nabla}_{\phi}F(\phi^{t},H^{t+1})\|^2\Big).
\end{align*}
Thus, the recursion from $s=0$ to $\tau_{\phi}-1$ generates
\begin{align*}
    \frac{1}{N}\sum_{i=1}^{N}\mathbb{E}_t\norm{{\phi}_i^{t,s}-\phi^t}^2 &\leq\sum_{s=0}^{\tau_{\phi}-1}(1+\frac{1}{2\tau_{\phi}-1})^s\Big[(1+\frac{1}{2\tau_{\phi}-1})6\eta_{\phi}^2(2\sigma_{\phi}^2+G^2)+\mathcal{T}_{2,\phi}^{\prime\prime\prime}\Big]\\
    &\leq(2\tau_{\phi}-1)\Big[(1+\frac{1}{2\tau_{\phi}-1})^{\tau_{\phi}-1}\Big] \Big[(1+\frac{1}{2\tau_{\phi}-1})6\eta_{\phi}^2(2\sigma_{\phi}^2+G^2)+\mathcal{T}_{2,\phi}^{\prime\prime\prime}\Big]\\    &\labelrel\leq{in_t2phi}3\tau_{\phi}\Big(\mathcal{T}_{2,\phi}^{\prime\prime\prime}+12\eta_{\phi}^2(2\sigma_{\phi}^2+G^2)\Big)\\    &\leq36\tau_{\phi}\eta_{\phi}^2(2\sigma_{\phi}^2+G^2)+18\eta_{\phi}^2\tau_{\phi}^2\Big(\sigma_{\phi}^2+\delta^2+\|{\nabla}_{\phi}F(\phi^{t},H^{t+1})\|^2\Big),
\end{align*}
where (\ref{in_t2phi}) holds due to $1+\frac{1}{2\tau_{\phi}-1}\leq2$ and $(1+\frac{1}{2\tau_{\phi}-1})^{\tau_{\phi}}\leq \sqrt{5}<\frac{5}{2}$ for any $\tau_{\phi}\geq1$.
\end{IEEEproof}
%%%%%%%%%%%%%%%%%%%%%%%%%%%%%
\subsection{Detailed Proof}
%%%%%%%%%%%%%%%%%%%%%%%%%%%%%
\begin{IEEEproof}
According to Lemmas~\ref{lemma_decomp},~\ref{lemma_t_1_phi},~\ref{lemma_t_2_phi},~\ref{lemma_t_h} and~\ref{lemma_t_3_phi}, we have
\begin{align}
    \mathbb{E}_t&[F( {\phi}^{t+1}, {H}^{t+1}) - F( {\phi}^{t}, {H}^{t+1})] \leq \eta_{\phi}\tau_{\phi}\tilde{\alpha}^tL_{\phi}^2q^2 -\frac{\bar{\alpha}^t\eta_{\phi}\tau_{\phi}}{2}\norm{\nabla_{\phi}F( {\phi}^{t}, \check{H}^{t+1})}^2 - \frac{\eta_{\phi}\bar{\alpha}^t}{2\tau_{\phi}}\norm{\frac{1}{\eta_{\phi}\bar{\alpha}^tN}\sum_{i=1}^{N}\bar{\Delta}_i^t}^2\notag\\
    &\quad+{L_{\phi}^2\eta_{\phi}^2\tau_{\phi}\bar{\alpha}^t}\left[q^2+36\tau_{\phi}\eta_{\phi}^2(2\sigma_{\phi}^2+G^2)+18\eta_{\phi}^2\tau_{\phi}^2\Big(\sigma_{\phi}^2+\delta^2+\|{\nabla}_{\phi}F(\phi^{t},H^{t+1})\|^2\Big)\right]+ \frac{L_{\phi}\sigma^2C^2d_1^2}{r^2N^2}\notag\\
    &\quad+3L_{{\phi}}\eta_{\phi}^2\tau_{\phi}(\sigma_{\phi}^{2}+L_{\phi}^2q^2+G^2) + 8\eta_{h}^2\tau_h^2L_h\chi^2(1-r)\frac{1}{n}\sum_{i=1}^{N}\norm{\nabla_{h} F_i(\phi^t,h_i^t)}^2 + 4\chi^2\eta_h^2\tau_h^2L_h\sigma_h^2(1-r)\notag\\
    &\labelrel\leq{in_lr}  -\frac{\bar{\alpha}^t\eta_{\phi}\tau_{\phi}}{4}\norm{\nabla_{\phi}F( {\phi}^{t}, \check{H}^{t+1})}^2  + (36\eta_{\phi}^4\tau_{\phi}^2L_{\phi}^2\bar{\alpha}^t+3\eta_{\phi}^2\tau_{\phi}L_{\phi})G^2 + (\tilde{\alpha}^t\eta_{\phi}\tau_{\phi}L_{\phi}^2+L_{\phi}^2\eta_{\phi}^2\tau_{\phi}\bar{\alpha}^t+3L_{\phi}^3\eta_{\phi}^2\tau_{\phi})q^2\notag\\
    &\quad+ [L_{\phi}^2\eta_{\phi}^2\tau_{\phi}\bar{\alpha}^t(72\tau_{\phi}\eta_{\phi}^2+18\eta_{\phi}^2\tau_{\phi}^2)+3L_{\phi}\eta_{\phi}^2\tau_{\phi}]\sigma_{\phi}^{2}+ 4\chi^2\eta_h^2\tau_h^2L_h\sigma_h^2(1-r) + \frac{L_{\phi}\sigma^2C^2d_1^2}{r^2N^2}\notag\\
    &\quad+ 8\eta_{h}^2\tau_h^2L_h\chi^2(1-r)\frac{1}{n}\sum_{i=1}^{N}\norm{\nabla_{h} F_i(\phi^t,h_i^t)}^2,\label{in_phi}
\end{align}
where~(\ref{in_lr}) holds due to $\eta \leq (1/(72\tau_{\phi}L_{\phi}))^{-2/3}$.
Combining (\ref{in_phi}) and Lemma~\ref{lemma_t_h}, if $128\eta_{\phi}L_{\phi}\tau_{\phi}\chi^2(r-1)\leq1$, we have
\begin{align*}
    \mathbb{E}_t&[F( {\phi}^{t+1}, {H}^{t+1}) - F( {\phi}^{t}, {H}^{t})] \leq  -\frac{\bar{\alpha}^t\eta_{\phi}\tau_{\phi}}{4}\norm{\nabla_{\phi}F( {\phi}^{t}, \check{H}^{t+1})}^2  + (36\eta_{\phi}^4\tau_{\phi}^2L_{\phi}^2\bar{\alpha}^t+3\eta_{\phi}^2\tau_{\phi}L_{\phi})G^2 \\
    &+ (\tilde{\alpha}^t\eta_{\phi}\tau_{\phi}L_{\phi}^2+L_{\phi}^2\eta_{\phi}^2\tau_{\phi}\bar{\alpha}^t+3L_{\phi}^3\eta_{\phi}^2\tau_{\phi})q^2+ [L_{\phi}^2\eta_{\phi}^2\tau_{\phi}\bar{\alpha}^t(72\tau_{\phi}\eta_{\phi}^2+18\eta_{\phi}^2\tau_{\phi}^2)+3L_{\phi}\eta_{\phi}^2\tau_{\phi}]\sigma_{\phi}^{2}\\
    &+ \frac{L_{\phi}\sigma^2C^2d_1^2}{r^2N^2} -\frac{\eta_h\tau_hr}{16}\frac{1}{n}\sum_{i=1}^{N}\norm{\nabla_{h} F_i(\phi^t,h_i^t)}^2 + \frac{\eta_h^2\tau_h^2L_h\sigma_h^2r}{2} + 4\eta_h^3L_h\tau_h^2(\tau_h-1)\sigma_h^2r+ 4\chi^2\eta_h^2\tau_h^2L_h\sigma_h^2(1-r).
\end{align*}
Taking an unconditional expectation, summing it over $t=0$ to $T-1$ and rearranging, we get
\begin{align*}
    \frac{1}{T}&\sum_{t=1}^{T-1}(\frac{\bar{\alpha}^t\eta_{\phi}\tau_{\phi}}{8}\mathbb{E}_{t}\norm{\nabla_{\phi}F( {\phi}^{t}, \check{H}^{t+1})}^2+\frac{\eta_h\tau_hr}{16N}\mathbb{E}\sum_{i=1}^{N}\norm{\nabla_{h} F_i(\phi^t,h_i^t)}^2)\leq\frac{\Delta F_0}{T}\\
    &+ (36\eta_{\phi}^4\tau_{\phi}^2L_{\phi}^2\frac{1}{T}\sum_{t=0}^{T-1    }\bar{\alpha}^t+3\eta_{\phi}^2\tau_{\phi}L_{\phi})G^2 + (\frac{1}{T}\sum_{t=0}^{T-1    }\tilde{\alpha}^t\eta_{\phi}\tau_{\phi}L_{\phi}^2+L_{\phi}^2\eta_{\phi}^2\tau_{\phi}\frac{1}{T}\sum_{t=0}^{T-1    }\bar{\alpha}^t+3L_{\phi}^3\eta_{\phi}^2\tau_{\phi})q^2\\
    &+ [L_{\phi}^2\eta_{\phi}^2\tau_{\phi}\frac{1}{T}\sum_{t=0}^{T-1    }\bar{\alpha}^t(72\tau_{\phi}\eta_{\phi}^2+18\eta_{\phi}^2\tau_{\phi}^2)+3L_{\phi}\eta_{\phi}^2\tau_{\phi}]\sigma_{\phi}^{2}+ 4\chi^2\eta_h^2\tau_h^2L_h\sigma_h^2(1-r) + \frac{L_{\phi}\sigma^2C^2d_1^2}{r^2N^2}\\
    & + \frac{\eta_h^2\tau_h^2L_h\sigma_h^2r}{2} + 4\eta_h^3L_h\tau_h^2(\tau_h-1)\sigma_h^2r
\end{align*}
This is a bound in terms of the virtual iterates $\check{H}^{t+1}$. However, we wish to show a bound in terms of the actual iterate $H^{t}$.
Using Lemma~\ref{lemma_cau_sch_in} and Assumption~\ref{assum_smooth}, we have
\begin{align*}
    \mathbb{E}_t[\nabla_{\phi} F( {\phi}^{t}, {H}^{t})- \nabla_{\phi}F( {\phi}^{t}, \check{H}^{t+1})] & \leq \frac{1}{N}\sum_{i=1}^{N}\mathbb{E}_t\norm{\nabla_{\phi} F_i( {\phi}^{t}, {h}_i^{t})- \nabla_{\phi}F_i( {\phi}^{t}, \check{h}_i^{t+1})}^2\\
    &\leq\frac{\chi^2L_{\phi}L_{h}}{N}\sum_{i=1}^{N}\mathbb{E}_t\norm{ \check{h}_i^{t+1}- {h}_i^{t})}^2\\
    &\labelrel\leq{in_3}\frac{\chi^2L_{\phi}L_{h}}{N}\sum_{i=1}^{N}\Big(16\eta_h^2\tau_h^2\norm{\nabla_{h}F_i(\phi^t,h_i^t)}^2+8\eta_h^2\tau_h^2\sigma_h^2\Big)\\
    &=8\eta_h^2\tau_h^2\sigma_h^2\chi^2L_{\phi}L_{h}+16 \eta_h^2\tau_h^2\chi^2L_{\phi}L_{h}\frac{1}{N}\mathbb{E}\sum_{i=1}^{N}\norm{\nabla_{h} F_i(\phi^t,h_i^t)}^2,
\end{align*}
where (\ref{in_3}) holds due the Lemma 23 in~\cite{pillutla2022federated}. Using
\begin{equation*}
    \norm{\nabla_{\phi} F( {\phi}^{t}, {H}^{t})}^2\leq 2\norm{\nabla_{\phi} F( {\phi}^{t}, {H}^{t})-\nabla_{\phi} F( {\phi}^{t}, \check{H}^{t+1})}^2 + 2\norm{\nabla_{\phi} F( {\phi}^{t}, \check{H}^{t+1})}^2
\end{equation*}
we have
\begin{equation*}
    \mathbb{E}\norm{\nabla_{\phi}F( {\phi}^{t}, {H}^{t+1})}^2 \leq 2 \mathbb{E}\norm{\nabla_{\phi}F( {\phi}^{t}, \check{H}^{t+1})}^2+16\eta_h^2\tau_h^2\sigma_h^2\chi^2L_{\phi}L_{h} + 32\eta_h^2\tau_h^2\sigma_h^2\chi^2L_{\phi}L_{h}\frac{1}{N}\mathbb{E}\sum_{i=1}^{N}\norm{\nabla_{h} F_i(\phi^t,h_i^t)}^2.
\end{equation*}
Thus, when ${32\gamma^2\chi^2\alpha}\leq\frac{1}{2}$,  we have
\begin{align*}
    \frac{\bar{\alpha}^t\eta_{\phi}\tau_{\phi}}{16} \mathbb{E}&\norm{\nabla_{\phi}F( {\phi}^{t}, {H}^{t+1})}^2+ \frac{\eta_{h}\tau_{h}r}{32}\frac{1}{N}\mathbb{E}\sum_{i=1}^{N}\norm{\nabla_{h} F_i(\phi^t,h_i^t)}^2\leq  \frac{\bar{\alpha}^t\eta_{\phi}\tau_{\phi}}{8}\mathbb{E}_{t}\norm{\nabla_{\phi}F( {\phi}^{t}, \check{H}^{t+1})}^2\\
    &+\frac{\eta_h\tau_hr}{16N}\mathbb{E}\sum_{i=1}^{N}\norm{\nabla_{h} F_i(\phi^t,h_i^t)}^2)+\bar{\alpha}^t\eta_{\phi}\tau_{\phi}\eta_h^2\tau_h^2\sigma_h^2\chi^2L_{\phi}L_{h}.
\end{align*}
Then, we have
\begin{align*}
    \frac{1}{T}&\sum_{t=1}^{T-1}(\frac{\bar{\alpha}^t\eta_{\phi}\tau_{\phi}}{8}\mathbb{E}_{t}\norm{\nabla_{\phi}F( {\phi}^{t}, {H}^{t+1})}^2+\frac{\eta_h\tau_hr}{16N}\mathbb{E}\sum_{i=1}^{N}\norm{\nabla_{h} F_i(\phi^t,h_i^t)}^2)\leq\frac{\Delta F_0}{T}\\
    &+ (36\eta_{\phi}^4\tau_{\phi}^2L_{\phi}^2\frac{1}{T}\sum_{t=0}^{T-1    }\bar{\alpha}^t+3\eta_{\phi}^2\tau_{\phi}L_{\phi})G^2 + (\frac{1}{T}\sum_{t=0}^{T-1    }\tilde{\alpha}^t\eta_{\phi}\tau_{\phi}L_{\phi}^2+L_{\phi}^2\eta_{\phi}^2\tau_{\phi}\frac{1}{T}\sum_{t=0}^{T-1    }\bar{\alpha}^t+3L_{\phi}^3\eta_{\phi}^2\tau_{\phi})q^2\\
    &+ [L_{\phi}^2\eta_{\phi}^2\tau_{\phi}\frac{1}{T}\sum_{t=0}^{T-1    }\bar{\alpha}^t(72\tau_{\phi}\eta_{\phi}^2+18\eta_{\phi}^2\tau_{\phi}^2)+3L_{\phi}\eta_{\phi}^2\tau_{\phi}]\sigma_{\phi}^{2}+ 4\chi^2\eta_h^2\tau_h^2L_h\sigma_h^2(1-r) + \frac{L_{\phi}\sigma^2C^2d_1^2}{r^2N^2}\\
    & + \frac{\eta_h^2\tau_h^2L_h\sigma_h^2r}{2} + 4\eta_h^3L_h\tau_h^2(\tau_h-1)\sigma_h^2r +\frac{1}{T}\sum_{t=0}^{T-1}\bar{\alpha}^t\eta_{\phi}\tau_{\phi}\eta_h^2\tau_h^2\sigma_h^2\chi^2L_{\phi}L_{h}
\end{align*}
Let $\eta_{\phi} = \mathcal{O}(1/(\tau_{\phi}L_{\phi}\sqrt{T}))$, $\eta_{h} = \mathcal{O}(1/(\tau_{h}L_{h}\sqrt{T}))$. As both $\frac{1}{T}\sum_{t=0}^{T-1}\tilde{\alpha}^t$ and $\frac{1}{T}\sum_{t=0}^{T-1}\bar{\alpha}^t$ are bounded, the big-$\mathcal{O}$ convergence about $T,$ we have
\begin{align*}
    \frac{1}{T}\sum_{t=0}^{T-1}(\frac{\bar{\alpha}^t}{L_{\phi}}\mathbb{E}\norm{\nabla_{\phi}F( {\phi}^{t}, {H}^{t+1})}^2&+\frac{r}{NL_{h}}\mathbb{E}\sum_{i=1}^{N}\norm{\nabla_{h} F_i(\phi^t,h_i^t)}^2) \leq \frac{\Delta F_0}{\sqrt{T}}+\mathcal{O}\big(\eta_{\phi}^3\frac{1}{T}\sum_{t=0}^{T-1}{\bar{\alpha}^t}(G^2+\sigma_{\phi}^2)\big)\\
    &+\mathcal{O}\big(\eta_{\phi}\frac{1}{T}\sum_{t=0}^{T-1}{\tilde{\alpha}^t}q^2\big) + \mathcal{O}\big(\eta_{h}^2\sigma_h^2\big) + \mathcal{O}(\frac{\sigma^2C^2d_1^2}{\eta_{\phi}r^2N^2}).
\end{align*}
\end{IEEEproof}

\end{document}